\documentclass{article} 
\usepackage{iclr2022_conference,times}

\usepackage{amsmath,amsfonts,bm}
\usepackage{multirow}
\usepackage{tabularx}
\usepackage{graphicx}
\usepackage{adjustbox}
\usepackage{amsthm}
\usepackage{MnSymbol}

\newcommand{\mset}[1]{\left\{\kern-.5em\left\{ #1 \right\}\kern-.5em\right\}}
\newcommand{\mmset}[1]{\{\kern-.4em\{ #1 \}\kern-.4em\}}

\newcommand{\vertiii}[1]{{\left\vert\kern-0.25ex\left\vert\kern-0.25ex\left\vert #1 
    \right\vert\kern-0.25ex\right\vert\kern-0.25ex\right\vert}}

\newcommand{\norm}[1]{\left\Vert#1\right\Vert}

\newcommand{\abs}[1]{\left\vert#1\right\vert}

\newcommand{\set}[1]{\left\{#1\right\}}
\newcommand{\parr}[1]{\left (#1\right )}
\newcommand{\brac}[1]{\left [#1\right ]}

\newcommand{\ip}[1]{\left \langle #1 \right \rangle }
\newcommand{\ipp}[1]{\left \llangle #1 \right \rrangle  }

\newcommand{\Real}{\mathbb R}

\newcommand{\eps}{\varepsilon}
\newcommand{\too}{\rightarrow}

\newcommand{\diag}{\textrm{diag}} 

\newcommand{\one}{\mathbf{1}}

\newcommand{\eg}{{e.g.}}
\newcommand{\ie}{{i.e.}}

 \newtheorem{theorem}{Theorem}
 
 \newtheorem{proposition}{Proposition}
 \newtheorem{corollary}{Corollary}
 \newtheorem{definition}{Definition}

\def\eqref#1{equation~\ref{#1}}

\def\1{\bm{1}}

\def\eps{{\epsilon}}

\def\vb{{\bm{b}}}

\def\ve{{\bm{e}}}

\def\vh{{\bm{h}}}

\def\vm{{\bm{m}}}
\def\vn{{\bm{n}}}

\def\vs{{\bm{s}}}
\def\vt{{\bm{t}}}
\def\vu{{\bm{u}}}
\def\vv{{\bm{v}}}

\def\vx{{\bm{x}}}

\def\vec1{{\bm{1}}}

\def\mA{{\bm{A}}}

\def\mC{{\bm{C}}}

\def\mI{{\bm{I}}}

\def\mL{{\bm{L}}}

\def\mO{{\bm{O}}}
\def\mP{{\bm{P}}}

\def\mR{{\bm{R}}}
\def\mS{{\bm{S}}}

\def\mW{{\bm{W}}}
\def\mX{{\bm{X}}}
\def\mY{{\bm{Y}}}
\def\mZ{{\bm{Z}}}

\DeclareMathAlphabet{\mathsfit}{\encodingdefault}{\sfdefault}{m}{sl}
\SetMathAlphabet{\mathsfit}{bold}{\encodingdefault}{\sfdefault}{bx}{n}
\newcommand{\tens}[1]{\bm{\mathsfit{#1}}}

\def\tX{{\tens{X}}}
\def\tY{{\tens{Y}}}

\def\gC{{\mathcal{C}}}

\def\gF{{\mathcal{F}}}

\def\gH{{\mathcal{H}}}

\def\gN{{\mathcal{N}}}

\def\gQ{{\mathcal{Q}}}

\def\gX{{\mathcal{X}}}

\newcommand{\R}{\mathbb{R}}

\usepackage[utf8]{inputenc} 
\usepackage[T1]{fontenc}    
\usepackage{hyperref}       
\usepackage{url}            
\usepackage{tabularx,booktabs}       
\usepackage{amsfonts}       
\usepackage{nicefrac}       
\usepackage{microtype, xcolor}      
\usepackage{wrapfig}
\usepackage{graphicx}
\usepackage{enumerate}  
\usepackage{multirow}
\usepackage{xfrac}

\title{Frame Averaging for Invariant and \\ Equivariant Network Design}

\author{Omri Puny\thanks{Equal contribution} $\, ^1$  \ \ Matan Atzmon\footnotemark[1] $\, ^1$ \ \ Heli Ben-Hamu\footnotemark[1] $\, ^1$  \ \  \bf{Ishan Misra}$^2$ \\ \bf{Aditya Grover}$^2$ \ \ Edward J. Smith$^2$ \ \ \bf{Yaron Lipman}$^2$ $^1$ \\
    $^1$Weizmann Institute of Science\ \
    $^2$Facebook AI Research}

\begin{document}

\maketitle
\vspace{-2pt}
\begin{abstract}
Many machine learning tasks involve learning functions that are known to be invariant or equivariant to certain symmetries of the input data. However, it is often challenging to design neural network architectures that respect these symmetries while being expressive and computationally efficient. For example, Euclidean motion invariant/equivariant graph or point cloud neural networks. \\
We introduce Frame Averaging (FA), a general purpose and systematic framework for adapting known (backbone) architectures to become invariant or equivariant to new symmetry types. Our framework builds on the well known group averaging operator that guarantees invariance or equivariance but is intractable. In contrast, we observe that for many important classes of symmetries, this operator can be replaced with an averaging operator over a small subset of the group elements, called a frame. We show that averaging over a frame guarantees exact invariance or equivariance while often being much simpler to compute than averaging over the entire group. Furthermore, we prove that FA-based models have maximal expressive power in a broad setting and in general preserve the expressive power of their backbone architectures. Using frame averaging, we propose a new class of universal Graph Neural Networks (GNNs), universal Euclidean motion invariant point cloud networks, and Euclidean motion invariant Message Passing (MP) GNNs. We demonstrate the practical effectiveness of FA on several applications including point cloud normal estimation, beyond $2$-WL graph separation, and $n$-body dynamics prediction, achieving state-of-the-art results in all of these benchmarks.

\end{abstract}

\section{Introduction}\vspace{-5pt}

Many tasks in machine learning (ML) require learning functions that are invariant or equivariant with respect to symmetric transformations of the data. For example, graph classification is invariant to a permutation of its nodes, while node prediction tasks are equivariant to node permutations. Consequently, it is important to design expressive neural network architectures that are by construction invariant or equivariant for scalable and efficient learning. This recipe has proven to be successful for many ML tasks including image classification and segmentation \citep{lecun1998gradient,long2015fully}, set and point-cloud learning \citep{zaheer2017deep,qi2017pointnet}, and graph learning \citep{kipf2016semi,gilmer2017neural,battaglia2018relational}.  

Nevertheless, for some important instances of symmetries, the design of invariant and/or equivariant networks is either illusive \citep{thomas2018tensor,dym2020universality}, computationally expensive or lacking in expressivity \citep{xu2018powerful,morris2019weisfeiler,maron2019provably,murphy2019relational}. In this paper, we propose a new general-purpose framework, called Frame Averaging (FA), that can systematically facilitate expressive invariant and equivariant networks with respect to a broad class of groups. At the heart of our framework, we build on a basic fact that arbitrary functions $\phi:V\too\Real$, $\Phi:V\too W$, where $V,W$ are some vector spaces, can be made invariant or equivariant by \emph{symmetrization}, that is averaging over the group \citep{yarotsky2021universal,murphy2018janossy}, \ie,  
\begin{equation}\label{e:group_averaging}
    \psi(X)=\frac{1}{|G|}\sum_{g\in G} \phi(g^{-1} \cdot X)  \quad \text{or} \quad  \Psi(X)=\frac{1}{|G|}\sum_{g\in G} g \cdot   \Phi(g^{-1} \cdot X).
\end{equation}
where $G=\set{g}$ denotes the group, $\psi:V\too\Real$ is invariant and $\Psi:V\too W$ is equivariant with respect to $G$.
Furthermore, since invariant and equivariant functions are fixed under group averaging, \ie, $\psi=\phi$ for invariant $\phi$ and $\Psi=\Phi$ for equivariant $\Phi$, the above scheme often leads to universal (\ie, maximally expressive) models \citep{yarotsky2021universal}. 
However, the challenge with \eqref{e:group_averaging} is that when the cardinality of $G$ is large (\eg, combinatorial groups such as permutations) or infinite (\eg, continuous groups such as rotations), then exact averaging is intractable. 
In such cases, we are forced to approximate the sum via heuristics or Monte Carlo (MC), thereby sacrificing the exact invariance/equivariance property for computational efficiency, \eg, \citet{murphy2018janossy,murphy2019relational} define heuristic averaging strategies for approximate permutation invariance in GNNs; similarly, \citet{hu2021forcenet} and \citet{shuaibi2021rotation} use MC averaging for approximate rotation equivariance in GNNs. 
A concurrent approach is to find cases where computing the symmetrization operator can be done more efficiently \citep{sannai2021equivariant}. 

The key observation of the current  paper is that the group average in \eqref{e:group_averaging} can be replaced with an average over a carefully selected subset $\gF(X)\subset G$ while retaining both exact invariance/equivariance and expressive power. Therefore, if $\gF$ can be chosen so that the cardinality $|\gF(X)|$ is mostly small, averaging over $\gF(X)$ results in both expressive and efficient invariant/equivariant model. We call the set-valued function $\gF:V\too 2^{G}$, a \emph{frame}, and show that it can successfully replace full group averaging if it satisfies a \textit{set equivariance property}. We name this framework Frame Averaging (FA) and it serves as the basis for the design of invariant/equivariant networks in this paper. 

We instantiate the FA framework by considering different choices of symmetry groups $G$, their actions on data spaces $V,W$ (manifested by choices of group representations), and the backbone architectures (or part thereof) $\phi,\Phi$ we want to make invariant/equivariant to $G$. We consider: (i) Multi-Layer Perceptrons (MLP), and Graph Neural Networks (GNNs) with node identification \citep{murphy2019relational,Loukas2020What} adapted to permutation invariant Graph Neural Networks (GNNs);  
(ii) Message-Passing GNN  \citep{gilmer2017neural} adapted to be invariant/equivariant to Euclidean motions, $E(d)$; (iii) Set network, DeepSets and PointNet \citep{zaheer2017deep,qi2017pointnet} adapted to be equivariant or \emph{locally} equivariant to $E(d)$; (iv) Point cloud network, DGCNN~\citep{wang2018dynamic}, adapted to be equivariant to $E(d)$.

Theoretically, we prove that the FA framework maintains the expressive power of its original backbone architecture which leads to some interesting corollaries: First, (i) results in invariant universal graph learning models; (ii) is an $E(d)$ invariant/equivariant GNN that maintain the power of message passing \citep{xu2018powerful,morris2019weisfeiler}; and (iii), (iv) furnish a universal permutation and $E(d)$ invariant/equivariant models. We note that both the construction and the proofs are arguably considerably simpler than the existing alternative constructions and proofs for this type of symmetry \citep{thomas2018tensor,fuchs2020se,dym2020universality}. We experimented with FA on different tasks involving symmetries including: point-cloud normal estimation, beyond $2$-Weisfeiler-Lehman graph separation, and $n$-body dynamics predictions, reaching state of the art performance in all. \vspace{-7pt}



\section{Frame Averaging}\vspace{-7pt}
In this section we introduce the FA approach using a generic formulation; in the next section we instantiate FA to different problems of interest. 

\vspace{-5pt}
\subsection{Frame averaging for function symmetrization}
%
Let $\phi:V\too\Real$ and $\Phi:V\too W$ be some arbitrary functions, where $V,W$ are normed linear spaces with norms $\norm{\cdot}_V,\norm{\cdot}_W$, respectively. For example, $\phi,\Phi$ can be thought of as neural networks. We consider a group $G=\set{g}$ that describes some symmetry we want to incorporate into $\phi, \Phi$. The way the symmetries $g\in G$ are applied to vectors in $V,W$ is described by the group's \emph{representations} $\rho_1:G\too \mathrm{GL}(V)$, and $\rho_2:G\too \mathrm{GL}(W)$, where $\mathrm{GL}(V)$ is the space of invertible linear maps $V\too V$ (automorphisms). A representation $\rho_i$ preserves the group structure by satisfying $\rho_i(gh)=\rho_i(g)\rho_i(h)$ for all $g,h\in G$ (see \eg, \cite{fulton2013representation}). As customary, we will sometimes refer to the linear spaces $V,W$ as representations.

Our goal is to make $\phi$ into an \emph{invariant} function, namely satisfy $\phi(\rho_1(g)X)=\phi(X)$, for all $g\in G$ and $X\in V$; and $\Phi$ into an \emph{equivariant} function, namely $\Phi(\rho_1(g)X)=\rho_2(g)\Phi(X)$, for all $g\in G$ and $X\in V$. We will do that by averaging over group elements, but instead of averaging over the entire group every time (as in \eqref{e:group_averaging}) we will average on a subset of the group elements called a \emph{frame}.

\pagebreak
\begin{definition} \label{def:frame}
A \emph{frame} is defined as a set valued function $\gF:V \too 2^{G} \setminus \emptyset$.
\begin{enumerate}
    \item A frame is $G$-\emph{equivariant} if $ \gF(\rho_1(g)X)= g \gF(X), \quad \forall X\in V,\  g\in G$, where as usual, $g\gF(X)=\set{g h \ \vert \ h\in \gF(X)}$, and the equality should be understood as equality of sets.
%
\item  A frame is \emph{bounded} over a domain $K\subset V$ if there exists a constant $c>0$ so that $\norm{\rho_2(g)}_{\mathrm{op}}\leq c$, for all $g\in \gF(X)$ and all $X\in K$, where $\norm{\cdot}_{\mathrm{op}}$ denotes the induced operator norm over $W$. 
\end{enumerate}
\end{definition}

\begin{wrapfigure}[7]{r}{0.37\textwidth}
  \begin{center}
  \vspace{-15pt}
    \includegraphics[width=0.35\textwidth]{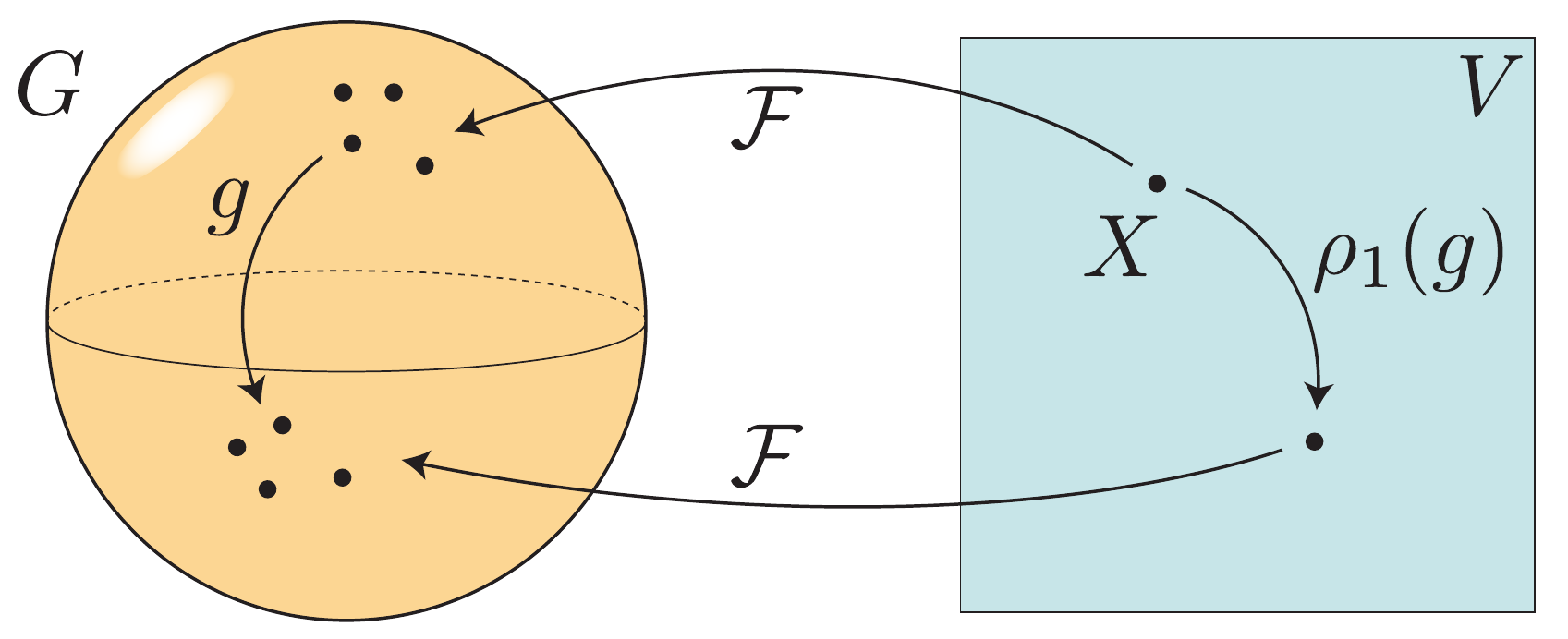}
  \end{center}
  \caption{Frame equivariance (sphere shape represents the group $G$; square represents $V$).}\label{fig:frame_equi}
\end{wrapfigure}
Figure \ref{fig:frame_equi} provides an illustration. 
How are equivariant frames useful?  Consider a scenario where an equivariant frame is easy to compute, and furthermore its cardinality, $|\gF(X)|$, is not too large. Then averaging over the frame, denoted $\ip{\cdot}_\gF$ and defined by
\begin{align}
    \label{e:inv}
    \ip{\phi}_\gF(X) &= \frac{1}{|\gF(X)|}\sum_{g\in \gF(X)} \phi(\rho_1(g)^{-1}X) \\
    \label{e:equi}
    \ip{\Phi}_\gF(X) &= \frac{1}{|\gF(X)|}\sum_{g \in \gF(X)} \rho_2(g) \Phi(\rho_1(g)^{-1}X) 
\end{align}
provides the required function symmetrization. In Appendix \ref{proof:frame} we prove:
\begin{theorem}[Frame Averaging]\label{thm:frame}
Let $\gF$ be a $G$ equivariant frame, and $\phi:V\too\Real$, $\Phi:V\too W$ some functions. Then, $\ip{\phi}_\gF$ is $G$ invariant, while $\ip{\Phi}_\gF$ is $G$ equivariant. 
\end{theorem}

Several comments are in order: First, the invariant case (\eqref{e:inv}) is a particular case of the equivariant case (\eqref{e:equi}) under the choice of $W=\Real$ and the trivial representation $\rho_2(g)\equiv 1$.
Second, in this paper we only consider $X$ and frame choices $\gF$ for which $\gF(X)$ are finite sets. Nevertheless, treating the infinite case is an important future research direction. 
Third, a trivial choice of an equivariant frame is $\gF(X)\equiv G$, that is, taking the frame to be the entire group for all $X\in V$ (for infinite but compact $G$ the sum in the FA in this case can be replaced with Harr integral). This choice can be readily checked to be equivariant, and turns the FA equations \ref{e:inv}, \ref{e:equi} into standard group averaging operators, \eqref{e:group_averaging}. The problem with this choice, however, is that it often results in an intractable or challenging computation, \eg, when the group is large or infinite. In contrast, as we show below, in some useful cases one can compute a manageable size frame and can use it to build invariant or equivariant operators in a principled way. Let us provide a simple example for Frame Averaging: consider $V=\Real^n$, $W=\Real$, and $G=\Real$ with addition as the group action. We choose the group actions\footnote{Note that since these are affine maps they are technically not representations but have an equivalent representation using homogeneous coordinates. 
Therefore, FA is also valid with affine actions as used here.}
in this case to be $\rho_1(a)\vx = \vx+a\one$, and $\rho_2(a)b=b+a$,  where $a,b\in \Real$, $\vx\in\Real^n$, and $\one\in\Real^n$ is the vector of all ones. We can define the frame in this case using the averaging operator $\gF(\vx) = \set{ \frac{1}{n}\one^T\vx} \subset G=\Real$. Note that in this case the frame contains only one element from the group, in other cases finding such a small frame is hard or even impossible. One can check that this frame is equivariant per Definition \ref{def:frame}. The FA of $\phi:\Real^n\too \Real$ would be $\ip{\phi}_\gF(\vx) = \phi(\vx - \frac{1}{n}\parr{\one^T\vx}\one)$ in the invariant case, and $\ip{\phi}_\gF(\vx) = \phi(\vx - \frac{1}{n}\parr{\one^T\vx}\one) + \frac{1}{n}\one^T\vx$ in the equivariant case.   

\paragraph{Incorporating $G$ as a second symmetry.} An important use case of frame averaging is with the backbones $\phi,\Phi$ already invariant/equivariant w.r.t.~some symmetry group $H$ and our goal is to make it invariant/equivariant to $H\times G$. For example, say we want to add $G=E(3)$ equivariance to permutation invariant set or graph functions, \ie, $H=S_n$.
We will provide sufficient conditions for the FA to provide this desired invariance/equivariance. First, let us assume $H$ is acting on $V$ and $W$ by the representations $\tau_1:H\too \mathrm{GL}(V)$ and $\tau_2:H\too \mathrm{GL}(W)$, respectively. Assume $\phi$ is $H$ invariant and $\Phi$ is $H$ equivariant. We say that representations $\rho_1$ and $\tau_1$ \emph{commute} if $\rho_1(g)\tau_1(h) X = \tau_1(h) \rho_1(g) X$ for all $g\in G$, $h\in H$, and $X\in V$. If $\rho_1$ and $\tau_1$ commute then the map $\gamma_1:H\times G \too \mathrm{GL}(V)$ defined by $\gamma_1(h,g) = \tau_1(h)\rho_1(g)$ is a representation of the group $H\times G$. Second, we would need that the frame $\gF(X)$ is \emph{invariant} to $H$, that is $\gF(\tau_1(h)X)=\gF(X)$. We show a generalization of Theorem~\ref{thm:frame}:
\begin{theorem}[Frame Average second symmetry]\label{thm:second_symmetry}
Assume $\gF$ is $H$-invariant and $G$-equivariant. Then,
\begin{enumerate}
\item 
If $\phi:V\too \Real$ is $H$ invariant and $\rho_1,\tau_1$ commute then $\ip{\phi}_\gF$ is $G\times H$ invariant. 
\item
If $\Phi:V\too W$ is $H$ equivariant and $\rho_i,\tau_i$,  $i={\scriptstyle 1,2}$, commmute then $\ip{\Phi}_\gF$ is $G\times H$~equivariant. 
\end{enumerate}
\end{theorem}

\paragraph{Right actions.} Above we used left actions for the definition of equivariance. There are other flavors of equivariance, \eg, if one of the actions is right. For example, if $g$ multiplies $\gF(X)$ from the right, then equivariance will take the form:
\begin{equation}\label{e:frame_right}
    \gF(\rho_1(g)X)=\gF(X) g^{-1}, \quad \forall X\in V,\  g\in G
\end{equation}
and accordingly
\begin{equation}
    \ip{\phi}_\gF(X) = \frac{1}{|\gF(X)|}\sum_{g\in \gF(X)} \phi(\rho_1(g)X), \ \ \ \ip{\Phi}_\gF(X) = \frac{1}{|\gF(X)|}\sum_{g\in \gF(X)}\rho_2(g)^{-1} \Phi(\rho_1(g)X) 
\end{equation}
are $G$ invariant and equivariant, respectively.

\paragraph{Efficient calculation of invariant frame averaging.}
There could be instances of the FA framework (indeed we discuss such a case later) where $|\gF(X)|$ is still too large to evaluate equations \ref{e:inv},\ref{e:equi}. In the invariant case, there is a more efficient form of FA, that can potentially be applied. To show it, let us start by defining the subgroup of symmetries of $X$, \ie, its stabilizer. The stabilizer of an element $X\in V$ is a subgroup of $G$ defined by $G_X=\set{g\in G \ \vert \ \rho_1(g)X=X}$. $G_X$ naturally induces an equivalence relation $\sim$ on $\gF(X)$, with $g \sim h \iff hg^{-1} \in G_X$. The equivalence classes (orbits) are $[g] = \left\{h \in \gF(X) \vert g \sim h \right\}=G_X g \subset \gF(X)$, for $g\in \gF(X)$, and the quotient set is denoted $\gF(X)/G_X$. 
\begin{theorem}\label{thm:orbits_decomp_and_equality}
Equivariant frame $\gF(X)$ is a disjoint union of equal size orbits, $[g]\in \gF(X)/G_X$.
\end{theorem}
The proof is in \ref{a:orbits_decomp_and_equality}. The first immediate consequence of Theorem \ref{thm:orbits_decomp_and_equality} is that the cardinality of $\gF(X)$ is at-least that of the stabilizer (intuitively, the inner-symmetries) of $X$, namely $|\gF(X)|\geq |G_X|$. Therefore, there could be cases, such as when $X$ describes a symmetric graph, where $|\gF(X)|$ could be too large to average over. A remedy comes from the following observation: for every $h\in [g]$, we have that $h=rg$, $r\in G_X$, and $\phi(\rho_1(h)^{-1}X) = \phi(\rho_1(g)^{-1}\rho_1(r)^{-1}X)=\phi(\rho_1(g)^{-1}X)$, since also $r^{-1}\in G_X$. Therefore the summands in equations \ref{e:inv} are constant over orbits, and we get 
\begin{equation}\label{e:qoutient_FA}
    \ip{\phi}_{\gF}(X)=\frac{1}{m_{_\gF}}\sum_{[g]\in {\gF(X)}/{G_X}} \phi(\rho_1(g)^{-1} X), 
\end{equation}
where $m_{_\gF}=\abs{\gF(X)/G_X}=\abs{\gF(X)} / \abs{G_X}$. This representation of invariant FA requires only $m_{_\gF}=|\gF(X)|/|G_X|$ evaluations, compared to $\vert \gF(X) \vert$ in the original FA in \eqref{e:inv}.\vspace{-5pt}
\paragraph{Approximation of invariant frame averaging.}
Unfortunately, enumerating $\gF(X)/G_X$ could be challenging in some cases. Nevertheless, \eqref{e:qoutient_FA} is still very useful: it turns out we can easily \emph{draw a random element} from $\gF(X)/G_X$ with uniform probability. This is an immediate application of the equal orbit size in Theorem \ref{thm:orbits_decomp_and_equality}: %
\begin{corollary}\label{cor:prob}
Let $\gF(X)$ be an equivariant frame, and $g\in \gF(X)$ be a uniform random sample. Then $[g]\in \gF(X)/G_X$ is also uniform.
\end{corollary}
Therefore, an efficient approximation strategy is averaging over uniform samples, $g_i\in\gF(X)$, $i\in[k]$,
\begin{equation}\label{e:FA_approx}
    \ipp{\phi}_\gF(X)=\frac{1}{k}\sum_{i=1}^k \phi(\rho_1(g_i)^{-1}X). 
\end{equation}
This approximation is especially useful, compared to the full-blown FA, when $m_{_\gF}=|\gF(X)|/|G_X|$ is small, \ie, when $|G_X|$ is large, or $X$ has many symmetries. Intuitively, the smaller $m_\gF$ the better the approximation in \eqref{e:FA_approx}. A partial explanation to this phenomenon is given in Appendix \ref{ss: role of m_F}, while an empirical validation is provided in Section \ref{ss:expressive power experiment}.
\vspace{-5pt}

\subsection{Expressive power}\vspace{-5pt}
\label{ss:expressive_power}
Another benefit in frame averaging as presented in equations \ref{e:inv} and \ref{e:equi} is that it preserves the expressive power of the base models, $\phi$, $\Phi$, as exaplined next. 
%
Consider some hypothesis function space $\gH=\set{\Phi}\subset \gC(V,W)$, where $\gC(V,W)$ is the set of all continuous functions $V\too W$. As mentioned above, the case of scalar functions $\phi$ is a special case where $W=\Real$, and $\rho_2(g)\equiv 1$. $\gH$ can be seen as the collection of all functions represented by a certain class of neural networks, \eg, Multilayer Perceptron (MLP), or DeepSets \citep{zaheer2017deep}, or Message Passing Graph Neural Networks \citep{gilmer2017neural}. 
We denote by $\ip{\gH}$ the collection of functions $\Phi\in \gH$ after applying the frame averaging in \eqref{e:equi},
    $\ip{\gH} = \set{ \ip{\Phi}_\gF \ \vert \ \Phi\in \gH}. $

We set some domain $K\subset V$ over which we would like to test the approximation power of $\ip{\gH}$. To make sure that FA is well defined over $K$ we will assume it is \emph{frame-finite}, \ie, for every $X\in K$, $\gF(X)$ is a finite set. Next, we denote $K_\gF=\set{\rho_1(g)^{-1}X \ \vert \ X\in K, g\in \gF(X)}$; intuitively, $K_\gF\subset V$ contains all the points sampled by the FA operator. Lastly, to quantify approximation error over a set $A\subset V$ let us use the maximum norm $\norm{\Phi}_{A,W}=\max_{X\in A}\norm{\Phi(X)}_W$. 
We prove that an arbitrary equivariant function $\Psi \in \gC(V,W)$ approximable by a function from $\gH$ over $K_\gF$ is approximable by an equivariant function from $\ip{\gH}$ (Proof details are found on Appendix \ref{proof:maximally_expressive}).

\begin{theorem}[Expressive power of FA]\label{thm:maximally_expressive} If $\gF$ is a bounded $G$-equivariant frame, defined over a frame-finite domain $K$, then for an arbitrary equivariant function $\Psi\in \gC(V,W)$ we have \begin{equation*}
    \inf_{\Phi\in \gH} \norm{\Psi-\ip{\Phi}_\gF}_{K,W} \leq c \inf_{\Phi\in \gH} \norm{\Psi-\Phi}_{K_\gF,W},
\end{equation*} 
Where $c$ is the constant from Definition \ref{def:frame}. \vspace{-5pt} 
\end{theorem} 
This theorem can be used to prove universality results if the backbone model is universal, even for non-compact groups (\eg, the Euclidean motion group). Below we will use it to prove universality of frame averaged architectures for graphs, and point clouds with Euclidean motion symmetry.   
\vspace{-5pt}

\section{Model instances}\vspace{-5pt}
\label{s:models}

We instantiate the FA framework by specifying: i) The symmetry group $G$, representations $\rho_1$, $\rho_2$ and the underlying frame $\gF$; ii) The backbone architecture for $\phi$ (invariant) or $\Phi$ (equivariant). \vspace{-6pt}

\subsection{Point clouds, Euclidean motions.} \vspace{-6pt}
\label{ss:point_clouds}
\textbf{Symmetry.} We would like to incorporate Euclidean symmetry to existing permutation invariant/equivaraint point cloud networks. The symmetry of interest is $G=E(d)=O(d)\ltimes T(d)$, namely the group of Euclidean motions in $\Real^d$ defined by rotations and reflections $O(d)$, and translations $T(d)$. We also discuss $G=SE(d)=SO(d)\ltimes T(d)$, where $SO(d)$ is the group of rotations in $\Real^d$. We define $V=\Real^{n\times d}$, and the group representation\footnote{Technically, this representation is defined by matrices $\tiny \begin{pmatrix}
\mR & \vt \\
\mathbf{0}^T & 1 
\end{pmatrix}$ acting on $\mX'=[\mX,\one]\in\Real^{n\times(d+1)}$. } is $\rho_1(g) \mX = \mX\mR^T + \one\vt^T$, where $\mR\in O(d)$ or $\mR\in SO(d)$, and $\vt\in\Real^d$ denotes the translation. $W,\rho_2$ are defined similarly, unless translation invariance is desired in which case we use the representation $\rho_2(g)\mX=\mX\mR^T$. 

\textbf{Frame.}
We define the frame $\gF(\mX)$ in this case based on Principle Component Analysis (PCA), as follows. Let $\vt=\frac{1}{n}\mX^T\one \in \Real^d$ be the centroid of $\mX$, and $\mC=(\mX-\one\vt^T)^T(\mX-\one\vt^T)\in\Real^{d\times d}$ the covariance matrix computed after removing the centroid from $\mX$. In the generic case the eigenvalues of $\mC$ satisfy $\lambda_1 < \lambda_2 < \cdots < \lambda_d$. Let $\vv_1,\vv_2,\ldots,\vv_d$ be the unit length corresponding eigenvectors. Then we define $\gF(\mX)=\set{(\brac{\alpha_1\vv_1,\ldots,\alpha_d\vv_d},t)\ \vert \  \alpha_i\in\set{-1,1} }\subset E(d)$. The size of this frame (when it is defined) is $2^d$ which for typical dimensions $d=2,3$ amounts to frames of size $4,8$, respectively. For $G=SE(d)$, we restrict $\gF(\mX)$ to orthogonal, positive orientation matrices; generically there are $2^{d-1}$ such elements, which amounts to $2, 4$ elements for $d=2,3$, respectively.
\begin{proposition}\label{prop:frame_E3_equivariance}
$\gF(\mX)$ based on the covariance and centroid are $E(d)$ equivariant and bounded. 
\end{proposition}
This choice of frame is defined for every $\mX\in V$ for which the covariance matrix $\mC$ has simple spectrum (\ie, non-repeating eigenvalues). It is known that within symmetric matrices, those with repeating eigenvalues are of co-dimension $2$ (see \eg, \cite{breiding2018geometry}). Therefore, $\gF(\mX)$ is defined for almost all $\gX$ except rare singular points. Where it is defined, $\gF$ is continuous as a direct consequence of perturbation theory of eigenvalues and eigenvectors of normal matrices (see \eg, Theorem 8.1.12 in \cite{golub1996matrix}). However, when very close to repeating eigenvalues, small perturbation can lead to large frame change. In Appendix \ref{s:frame_analysis} we present an empirical study of frame stability and likelihood of encountering repeating eigenvalues in practice.

Since we would like to incorporate $E(d)$ symmetries to an already $S_n$ invariant/equivariant architectures, per Theorem \ref{thm:second_symmetry}, we will also need to show that the $\rho_1$ (and similarly $\rho_2$) commute with $\tau:S_n\too \mathrm{GL}(V)$ defined by $\tau(h)\mX=\mP \mX$, where $\mP=\mP_h$, $h\in S_n$, is the permutation representation. That is $\mP_h\in\Real^{n\times n}$ is the permutation matrix representing $h\in S_n$, that is, $\mP_{ij} = 1$ if $i=h(j)$ and $0$ otherwise. Indeed $\tau(h)\rho_1(g)\mX= \mP(\mX\mR^T + \one\vt^T) = \rho_1(g)\tau(h)\mX$. Furthermore, note that $\gF(\tau(h)\mX)=\gF(\mX)$, therefore $\gF$ is also $S_n$ invariant. \vspace{5pt} \\
\textbf{Backbone architectures.}
We incorporate the FA framework with two existing popular point cloud network layers: i) PointNet \citep{qi2017pointnet}; and ii) DGCNN \citep{wang2018dynamic}. We denote both architectures by $\Phi_{d,d'}:\Real^{n\times d} \too \Real^{n\times d'}$ for the $S_n$ equivariant version of these models. To simplify the discussion, we omit particular choices of layers and feature dimensions, Appendix \ref{app:impl_dets_PC} provides the full details. We experimented with two design choices: First, consider frame averaged $\Phi_{3,3}$, \ie, $\ip{\Phi_{3,3}}_\gF$, yielding a universal, $E(3)$ equivariant versions for PointNet and DGCNN, dubbed FA-PointNet and FA-DGCNN. A more complex design choice, taking inspiration from deep architectures with multiple modules, is to compose blocks of FA networks. For example, to build a \emph{local} $E(3)$ equivariant version of PointNet, denote also  $\Upsilon_{d,d'}:\Real^{d}\too\Real^{d'}$  an MLP. We decompose the input point cloud to $k$-nn patches $\mX_i\in\Real^{k\times 3}$, $i\in [n]$, where each patch is sorted by distance. Next, feed each patch into an equivariant FA $\Upsilon_{3k,3d}$, each with its own frame $\gF_i$, resulting in  $E(3)$ equivariant features in $\Real^{3d}$ for every point, $\mY\in\Real^{n\times 3d}$; then applying equivariant FA $\Phi_{3d,3}$ providing output in $\Real^{n\times 3}$. That is, $\Psi(\mX)=\ip{\Phi_{3d,3}}_\gF\parr{\brac{\ip{\Upsilon_{3k,3d}}_{\gF_1}(\mX_1),\ldots,\ip{\Upsilon_{3k,3d}}_{\gF_n}(\mX_n)}}$, where brackets denote concat in the first dimension. We name this construction FA-Local-PointNet.\vspace{5pt} \\
\textbf{Universality.}
We use Theorem \ref{thm:maximally_expressive} to prove that using a universal set invariant/equivariant backbone $\phi,\Phi$, such as DeepSets or PointNet (see \eg,  \cite{zaheer2017deep,qi2017pointnet,segol2019universal}) leads to a universal model. Let $\gH$ be any such universal set-equivariant function collection. That is, for arbitrary continuous set function $\Psi$ we have (in the notation of Section \ref{ss:expressive_power})  $\inf_{\Phi\in\gH}\norm{\Psi-\Phi}_{\Omega,W}=0$ for arbitrary compact sets $\Omega\subset V$.
If $K\subset V$ is some bounded domain, then the choice of frame $\gF$ described above implies that $K_\gF$ is also bounded and therefore contained in some compact set $\Omega\subset V$. Consequently, Proposition \ref{prop:frame_E3_equivariance} and Theorem \ref{thm:maximally_expressive} imply Corollary \ref{cor:fa_deepsets_universality}. A similar results holds for $SE(d)$, which provides a similar expressiveness guarantee as the one from \citep{dym2020universality} analyzing Tensor Field Networks \citep{thomas2018tensor,fuchs2020se}. 
\begin{corollary}[FA-DeepSets/PointNet are universal]\label{cor:fa_deepsets_universality}
Frame Averaging DeepSets/PointNet using the frame $\gF$ defined above results in a universal $E(d)\times S_n$ invariant/equivariant model over bounded frame-finite sets $K\subset V$. 
\end{corollary}

\vspace{-5pt}

\subsection{Graphs, permutations} 
\label{ss:graphs_perms}
\vspace{-5pt}
\textbf{Symmetry and frame.} Let $G=S_n$, and $V=\Real^{n\times d}\times  \Real^{n\times n}$, where $\tX=(\mY,\mA)\in V$ represents a set of node features $\mY\in\Real^{n\times d}$, and an adjacency matrix (or some edge attributes) $\mA\in\Real^{n\times n}$;  we assume undirected graphs, meaning $\mA=\mA^T$. Let $\tX\in V$, the representation $\rho_1$ is defined by $\rho_1(g)\tX = (\mP\mY, \mP \mA \mP^T)$, where $\mP=\mP_g$ is the permutation matrix representing $g\in S_n$.
We define $\gF(\tX)$ to contain all $g\in S_n$ that sort the rows of the matrix $\mS(\tX)$ in column lexicographic manner, \ie, $\mP \mS(\tX)$ is lexicographically sorted; the matrix $\mS$ is defined as follows.
%
Let $\mL=\diag(\mA\one)-\mA$ be the graph's Laplacian. For every eigenspace (traversed in increasing eigenvalue order), spanned by the orthogonal basis $\vu_1,\ldots,\vu_k$, we add the (equivariant) column $\diag(\sum_{i=1}^k\vu_i\vu_i^T)$ to $\mS$ \citep{furer2010power}. Hence, the number of columns of $\mS$ equals to the number of unique eigenvalues of $\mL$.
\begin{proposition}\label{prop:sorting_frame}
$\gF(\tX)$ defined by sorting of $\mS(\tX)$ is $S_n$-equivariant and bounded. 
\end{proposition}
\vspace{-2pt}
\textbf{Backbone architectures.} In this case we perform only invariant tasks, for which we chose two universal backbone architectures for $\phi$: (i) MLP applied to $\text{vec}(\mY,\mA)$ and (ii) GNN+ID \citep{murphy2019relational,Loukas2020What}, denoting a GNN backbone equipped with node identifiers as node features. We perform FA with $\phi$ according to the frame constructed in Proposition \ref{prop:sorting_frame}. Note that Theorem \ref{thm:orbits_decomp_and_equality} implies, $\abs{\gF(X)}\geq\abs{\mathrm{Aut}(G)}$, since the stabilizer $G_\tX$ is the automorphism group of the graph, \ie, $g\in G_\tX=\mathrm{Aut}(\tX)$ iff $\rho_1(g)\tX = \tX$.  This means that for symmetric graphs equation \ref{e:inv} can prove too costly. In this case, we use the approximate invariant FA, \eqref{e:FA_approx}.\vspace{5pt}\\
\textbf{Universality.}
Let us use Theorem \ref{thm:maximally_expressive} to prove FA-MLP and FA-GNN+ID are universal. Again, it is enough to consider the equivariant case. Let $\gH\subset \gC(V,W)$ denote the collection of functions that can be represented by MLPs or GNN+ID. Universality results in \citep{pinkus1999approximation,Loukas2020What,puny2020global} imply that for any continuous graph (\ie, $S_n$ equivariant) function $\Psi$, $\inf_{\Phi\in \gH}\norm{\Psi-\Phi}_{\Omega,W}=0$ for any compact $\Omega\subset V$. Let $K\subset V$ be some bounded domain, then since $G$ is finite then $K_\gF$ is also bounded and is contained is some compact set. Proposition \ref{prop:sorting_frame} and Theorem \ref{thm:maximally_expressive} now imply:
\begin{corollary}[FA-MLP and FA-GNN+ID is graph universal] 
Frame Averaging MLP/GNN+ID using the frame $\gF$ above results in a universal $S_n$ equivariant graph model over bounded domains $K\subset V$. 
\end{corollary}


\vspace{-5pt}

\subsection{Graphs, Euclidean
motions.} \vspace{-5pt}
\label{ss:graphs_perms_and_euclidean}

\paragraph{Symmetry and frame.} We consider the group $G=E(d)$ acting on graphs, \ie, $V=\Real^{n\times d} \times  \Real^{ n\times n}$, where $\tX=(\mY,\mA)\in V$ represents a set of node and edge attributes, as described above. The group representation is $\rho_1(g)\tX = \rho_1(g)(\mY,\mA) = (\mY\mR^T+\one\vt^T,\mA)$. We define the frame $\gF(\tX)=\gF(\mY)$ using the node features as in the point cloud case, Section \ref{ss:point_clouds}. Therefore Proposition \ref{prop:frame_E3_equivariance} implies $\gF$ is equivariant and bounded. Next, also in this case we would like to incorporate $E(d)$ symmetries to an already $S_n$ invariant/equivariant graph neural network architectures; again per Theorem \ref{thm:second_symmetry}, we will also need to show that the $\rho_1$ (and similarly $\rho_2$) commutes with $\tau:S_n\too \mathrm{GL}(V)$ defined by $\tau(h)\tX=(\mP \mY, \mP \mA \mP^T)$, where $\mP=\mP_h$ is the permutation matrix of $h\in S_n$. Indeed $\tau(h)\rho_1(g)\tX= (\mP( \mY\mR^T+\one\vt^T), \mP \mA \mP^T)=\rho_1(g)\tau(h)\tX$. Furthermore, note that as in the point cloud case $\gF(\tau(h)\tX)=\gF(\tX)$, therefore $\gF$ is also $S_n$ invariant.\vspace{5pt} \\ 
\textbf{Backbone architecture.}
The backbone architecture we chose for this instantiation is the Message Passing GNN in \cite{gilmer2017neural}, an $S_n$ equivariant model denoted $\Phi_{d,d'}:\Real^{n\times d}\times \Real^{n\times n}\too \Real^{n\times d'}\times \Real^{n\times n}$.
In this case we constructed a model, as suggested above, 
by composing 
$l$ equivariant layers $\Psi(\tX) = \langle\Phi^{{\scriptscriptstyle (l)}}_{3d',3}\rangle_{\gF}\circ ... \langle\Phi^{{\scriptscriptstyle (i)}}_{3d',3d'}\rangle_{\gF} ...\circ\langle \Phi^{{\scriptscriptstyle (1)}}_{6,3d'}\rangle_\gF(\tX)$.
The input feature size is $6$ since we use velocities in addition to initial position as input ($n$-body problem). We name this model FA-GNN.\vspace{-10pt}


\section{Previous works}\vspace{-5pt}

\paragraph{Rotation invariant and equivariant point networks.}
%
State of the art $S_n$ invariant networks, \eg,  \citep{qi2017pointnet,qi2017pointnet++,atzmon2018point,li2018pointcnn,xu2018spidercnn,wang2018dynamic} are not invariant/equivariant to rotations/reflections by construction \citep{chen2019clusternet}.
%
Invariance to global or local rotations can be achieved by  modifying the 3D convolution operator or modifying the input representation. %
Relative angles and distances across points~\citep{deng2018ppf,zhang2019rotation} or angles and distances w.r.t.~normals~\citep{gojcic2019perfect} can be used for rotation invariance.
%
Other works use some local or global frames to achieve invariance to rotations and translations.
\cite{xiao2020endowing,yu2020deep,deng2018ppf} also use PCA to define rotation invariance, and can be seen as instances of the FA framework. We augment this line of work by introducing a more general framework that includes equivariance to rotation/reflection and translation, more general architectures and symmetries as well as theoretical analysis of the expressive power of such models.\\
%
%
%
%
Equivariance is a desirable property for 3D recognition, registration~\citep{ao2021spinnet}, and other domains such as complex physical systems~\citep{kondor2018n}.
A popular line of work utilizes the theory of spherical harmonics to achieve equivariance ~\citep{worrall2017harmonic,esteves2018learning,liu2018deep,weiler20183d,cohen2018spherical}.
Notably, Tensor Field Networks (TFN), $SE(3)$ transformers, and Group-Equivariant Attention \citep{thomas2018tensor,fuchs2020se,romero2020group} achieve equivariance to both translation and rotation, \ie, $SE(3)$, and are maximally expressive (\ie, universal) as shown in~\cite{dym2020universality}. These methods, however, are specifically adapted to $SE(3)$ and require high order $SO(3)$ representations as features. 
%
Recently~\cite{deng2021vector} propose a rotation equivariant network by introducing tensor features, linear layers that act equivariantly on them and equivariant non-linearities etc. However, their architecture is not proved to be universal.
Discrete convolutions~\citep{cohen2019gauge,li2019discrete,worrall2018cubenet} have also been used for achieving equivariance.
In particular,~\cite{chen2021equivariant} propose point networks that are $SE(3)$ equivariant and use separable discrete convolutions.
Lastly, \citep{finzi2020generalizing} construct equivariant layers using local group convolution, and extends beyond rotations to any Lie group.\vspace{2pt} \\
\textbf{Graph neural networks.} 
%
Message passing (MP) GNNs~\citep{gilmer2017neural} are designed to be $S_n$ equivariant. \cite{kondor2018covariant} introduces a broader set of equivariant operators in MP-GNNs, while \cite{maron2018invariant} provides a full characterization of linear permutation invariant/equivariant GNN layers. 
In a parallel approach, trying to avoid harming expressively due to restricted architectures \citep{xu2018powerful,morris2019weisfeiler}, other works suggested symmetrization of non invariant/equivariant backbones. Ranging from eliminating all symmetries by a canonical ordering \citep{niepert16} to averaging over the entire symmetry group \citep{murphy2019relational,murphy2018janossy}, which amount to the trivial frame $\gF\equiv\rho(G)$, with the symmetry group $G=S_n$. As we have also shown, this approach comes at a cost of high variance approximations hindering the learning process. Our FA framework reduces the variance both by choosing a canonical ordering and addressing the fact that it may not be unique.    

Recently, a body of work studies GNNs with invariance/equivariance to $E(3)$ (or a similar group) to deal with symmetries in molecular data or dynamical systems . Many $SE(3)$ equivariant constructions~\citep{anderson2019cormorant,fuchs2020se,batzner2021se,klicpera2021gemnet} extend TFN~\citep{thomas2018tensor} and inherit its expensive higher order feature representations in the intermediate layers.
Finally, a recent work by \citet{satorras2021n} provides an efficient message passing construction which is $E(d)$ equivariance but is not shown to be universal, thus far.
\vspace{-5pt}









\section{Experiments}\vspace{-5pt}
We evaluate our FA framework on a few invariant/equivaraint point cloud and graph learning tasks: point cloud normal estimation ($O(3)$ equivariant and translation invariant); graph separation tasks ($S_n$ invariant); and particles position estimation, \ie, the $n$-body problem ($E(3)$ equivariant).  \vspace{-5pt}

\subsection{Point Clouds: Normal Estimation}\vspace{-5pt}
\label{ss:normal_estimation}
Normal estimation is a core \emph{geometry processing} task, where the goal is to estimate normal data from an input point cloud, $\mX\in\Real^{n\times 3}$. This task is $O(3)$ equivariant and translation invariant. To test the effect of rotated data on the different models we experimented with three different settings: i) $I/I$ - training and testing on the original data; ii) $I/SO(3)$ - training on the original data and testing on randomly rotated data; and iii) $SO(3)/SO(3)$ - training and testing with randomly rotated data. We used the ABC dataset~\citep{Koch_2019_CVPR} that contains 3 collections (10k, 50k, and 100k models each) of \emph{Computer-Aided Design} (CAD) models. We follow the protocol of the benchmark suggested in \cite{Koch_2019_CVPR}, and quantitatively measure normal estimation quality via  $1 - (\vn ^T \hat{\vn})^2$, with $\vn$ being the ground truth normal and $\hat{\vn}$ the normal prediction. We used the same random train/test splits from \citet{Koch_2019_CVPR}. For baselines, we chose the PointNet~\citep{qi2017pointnet} and DGCNN~\citep{wang2018dynamic} architectures, which are popular permutation equivariant 3D point cloud architectures. In addition, we also compare to VN-PointNet and VN-DGCNN~\citep{deng2021vector}, a recent state of the art approach for $SO(3)$ equivariant network design. We also tested our FA models, FA-PointNet and FA-DGCNN as described in Section \ref{ss:point_clouds}. In addition, we tested our local $O(3)$ equivariant model, FA-Local-PointNet. See Appendix \ref{app:impl_dets_PC} for the further implementation details. The results in Table \ref{tab:normal} showcase the advantages of our FA framework: Incorporating FA to existing architectures is beneficial in scenarios (ii-iii), outperforming augmentation by a large margin. In contrast to VN models, FA models maintain (and in some cases even improve) the baseline estimation quality (i), attributed to the expressive power of the FA models.

\begin{table}[h]
\centering
\resizebox{\textwidth}{!}{%
\begin{tabular}{l|lll|lll|lll}
Method & \multicolumn{3}{l}{10k} & \multicolumn{3}{l}{50k} & \multicolumn{3}{l}{100k} \\                & $I/I$ & $I/SO(3)$ & $SO(3)/SO(3)$ & $I/I$ & $I/SO(3)$ & $SO(3)/SO(3)$ & $I/I$ & $I/SO(3)$ & $SO(3)/SO(3)$
     \\ 
        \hline
PointNet  & .207$\pm$.004 & .449$\pm$.006 & .258$\pm$.002 & .188$\pm$.002 & .430$\pm$.007 & .232$\pm$.001 & .188$\pm$.006 & .419$\pm$.006 & .231$\pm$.001 \\
      
VN-PointNet  & .215$\pm$.003 & .216$\pm$.004 & .223$\pm$.004  & .185$\pm$.002 & .186$\pm$.006 & .187$\pm$.006 & .189$\pm$.004 & .188$\pm$.007 & .185$\pm$.003 \\

FA-PointNet & .158$\pm$.001 & .163$\pm$.001 & .161$\pm$.002 & .148$\pm$.001 & .148$\pm$.003 & .150$\pm$.002 & .148$\pm$.002 & .147$\pm$.001& .149$\pm$.001 \\

FA-Local-PointNet  & .097$\pm$.001 & .098$\pm$.001 & .098$\pm$.001 & .091$\pm$.001 & .090$\pm$.001 & .091$\pm$.001 & .091$\pm$.001 & .090$\pm$.002 & .091$\pm$.002\\

DGCNN  & .070$\pm$.003& .193$\pm$.015 & .121$\pm$.001 & \textbf{.061$\pm$.004} & .174$\pm$.007 & .122$\pm$.001 & \textbf{.058$\pm$.002} & .173$\pm$.003 & .112$\pm$.001 \\

VN-DGCNN  & .133$\pm$.003 & .130$\pm$.001 & .144$\pm$.007 & .127$\pm$.005 & .125$\pm$.001 & .127$\pm$.006 & .127$\pm$.005 & .125$\pm$.001 & .127$\pm$.005 \\

FA-DGCNN & \textbf{.067$\pm$.001} & \textbf{.069$\pm$.002} & \textbf{.071$\pm$.003} & .065$\pm$.001 & \textbf{.067$\pm$.004} & \textbf{.068$\pm$.004} & .073$\pm$.008 & \textbf{.067$\pm$.001} & \textbf{.071$\pm$.009} \\

\vspace{-7pt}
\end{tabular}}
\caption{Normal estimation, ABC dataset \citep{Koch_2019_CVPR} benchmark.\vspace{-5pt}}
\label{tab:normal}
\end{table}

\begin{wraptable}[14]{r}{0.35\textwidth}
\renewcommand{\tabcolsep}{1.6pt}
\vspace{-26pt}
\begin{adjustbox}{width=0.35\textwidth}
\begin{tiny}
\begin{tabular}{lccc}
    \toprule
    Model   & \multicolumn{1}{l}{GRAPH8c} & \multicolumn{1}{l}{EXP} & \multicolumn{1}{l}{EXP-classify} \\ \midrule
    GCN     & 4755                        & 600                     & 50\%                             \\
    GAT     & 1828                        & 600                     & 50\%                             \\
    GIN     & 386                         & 600                     & 50\%                             \\
    CHEBNET & 44                      & 71                      & 82\%                             \\
    PPGN    & 0                      & 0                       & \textbf{100\%}                            \\
    GNNML3  & 0                      & 0                       &\textbf{ 100\% }                           \\ \hline
    GA-MLP  &                       0      &     0                    &      50\%                            \\
    FA-MLP  &                  0           &      0                   &   \textbf{100\%}                               \\ \hline
    GA-GIN+ID  &  0                           & 0                       & 50\%                               \\
    FA-GIN+ID  &  0                          & 0                       & \textbf{100\%}                            \\ \bottomrule
    \end{tabular}
\end{tiny}
\end{adjustbox}
\caption{Graph separation~\citep{balcilar2021breaking,ACGL-IJCAI21}.}
\label{tab:exp}
\end{wraptable}
\vspace{-5pt}
\subsection{Graphs: Expressive Power}
\label{ss:expressive power experiment}
Producing GNNs that are both expressive and computationally tractable is a long standing goal of the graph learning community. In this experiment we test graph separation ($S_n$ invariant task): the ability of models to separate and classify graphs, a basic trait for graph learning. We use two datasets: GRAPH8c~\citep{balcilar2021breaking} that consists of all non-isomorphic, connected $8$ node graphs; and EXP~\citep{ACGL-IJCAI21} that consists of 3-WL distinguishable graphs that are not 2-WL distinguishable. There are two tasks: (i) count pairs of graphs not separated by a randomly initialized model in GRAPH8c and EXP; and (ii) learn to classify EXP to two classes. We follow \cite{balcilar2021breaking} experimental setup. As baselines we use GCN~\cite{kipf2016semi}, GAT~\cite{Petar2018graph}, GIN~\cite{xu2018powerful}, CHEBNET~\cite{tang2019chebnet}, PPGN~\cite{maron2019provably}, and GNNML3~\cite{balcilar2021breaking}, all of which are equivariant by construction. We compare to our FA-MLP, and FA-GIN+ID, as described in Section \ref{ss:graphs_perms} that provide tractable option for universal GNNs. We also compare to the trivial (\ie, entire group) frame averaging $\gF\equiv G$, denoted GA-MLP and GA-GIN+ID as advocated in~\citep{murphy2019relational}. Appendix \ref{appendix:expressive power} provides full implementation details. 
%

\begin{wrapfigure}[13]{r}{0.5\textwidth}
\vspace{-10pt}
  \begin{center}
       \includegraphics[width=0.5\textwidth]{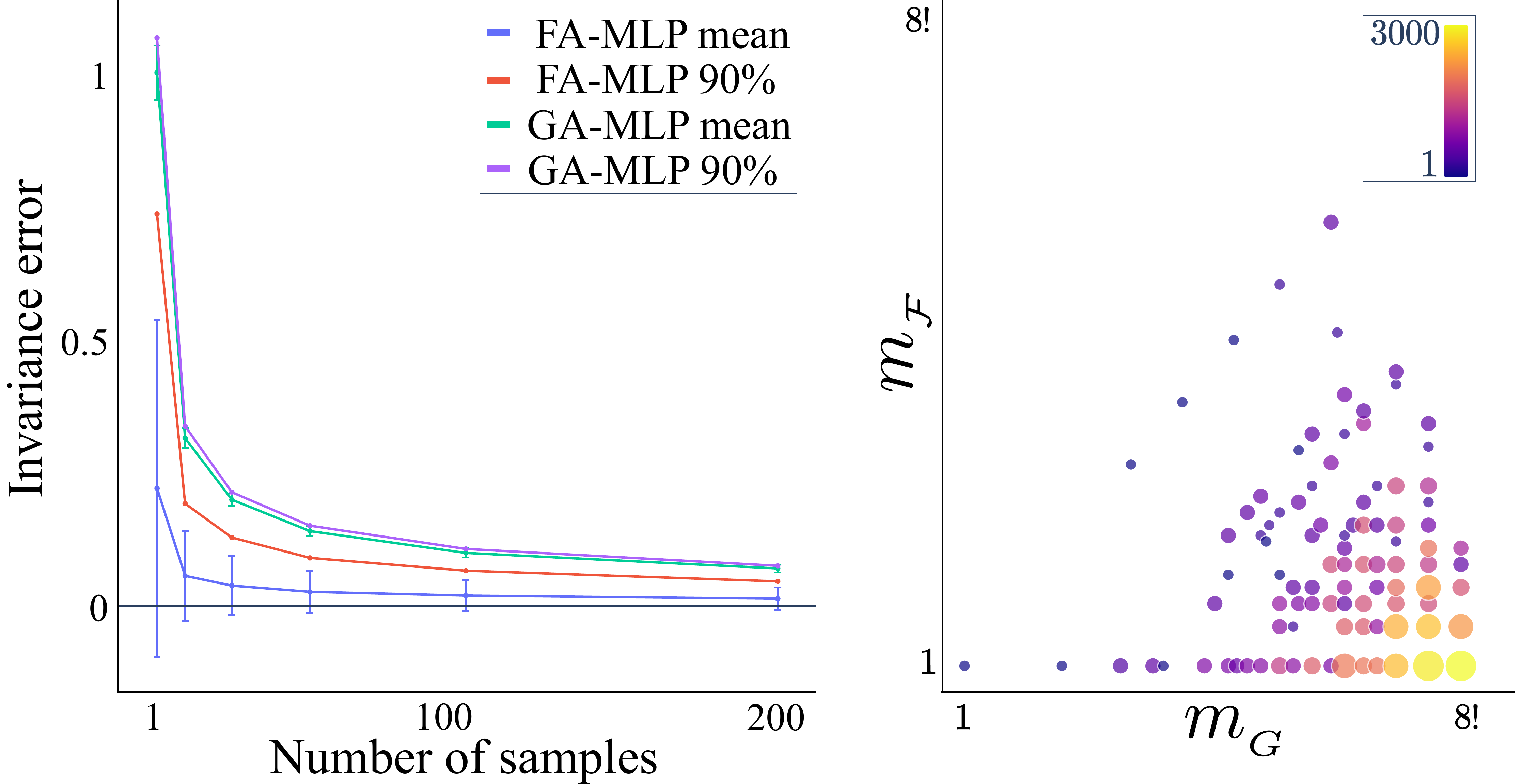}
  \end{center}
  \vspace{-10pt}
  \caption{Invariance error (left); $m_{_\gF}$-$m_{_G}$ (right).}\label{fig:inverr_m_f}
\end{wrapfigure}
The two first columns in Table \ref{tab:exp} show the results of the separation task (i). As expected both FA and GA provide perfect separation, however, are not perfectly invariant unless one computes the full averages on $\gF$ and $G$, respectively. Since for some graphs (with many symmetries) full averages are not feasible, we compare the \emph{invariance error} (see appendix for the exact definition) of GA with that of the approximate FA, \eqref{e:FA_approx}, for $50$ randomly permuted inputs. Figure \ref{fig:inverr_m_f}-left shows the mean, std, and $90\%$ percentile of \emph{invariance error} for increasing sample size $k$. Note that approximate FA is more invariant even with as little as $k=1$ samples, which explains why in the classification task (ii), in Table \ref{tab:exp}, right column, FA is able to learn while GA fails, when both are trained with sample size $k=1$. Figure \ref{fig:inverr_m_f}-right compares $m_\gF$ and $m_{G}$ (the maximal number of unique elements in the FA, \eqref{e:qoutient_FA}) for GRAPH8c dataset. The color and size of the points in the plot represents how many graphs in the dataset have the corresponding $(m_G,m_\gF)$ values.  This demonstrates the benefit (in one example) of FA over GA in view of Theorem \ref{thm:large_deviations} and approximation in \eqref{e:FA_approx}. Of course, other, more powerful frames can be chosen, \eg, using higher order WL \citep{morris2019weisfeiler} or substructure counting \citep{bouritsas2020improving} that will further improve the approximation of \eqref{e:FA_approx}.\vspace{-5pt}

\subsection{Graphs: $n$-Body Problem }

\label{ss:n-body}
\begin{wraptable}[12]{r}{0.35\textwidth}
\renewcommand{\tabcolsep}{1.6pt}
\vspace{-10pt}
\begin{adjustbox}{width=0.35\textwidth}
\begin{tiny}
\begin{tabular}{lcc}
\hline
Method            & MSE    & Forward time (s) \\ \hline
Linear            & 0.0819 & .0001   \\
SE(3) Transformer & 0.0244 & .1346  \\
TFN               & 0.0155 & .0343   \\
GNN               & 0.0107 & .0032   \\
Radial Field      & 0.0104 & .0039    \\
EGNN              & 0.0071 & .0062    \\
FA-GNN              & \textbf{0.0057} & .0041                  \\ \hline
\end{tabular}
\end{tiny}
\end{adjustbox}
\caption{$n$-body experiment \citep{satorras2021n}. }
\label{tab:n_body}
\end{wraptable}
In this task we learn to solve the $n$-body problem ($E(3)$ equivariant). The dataset, created in \citep{satorras2021n,fuchs2020se}, consists of a collection of $n=5$ particles systems, where each particle is equipped with its initial position in $\Real^3$ and velocity in $\Real^3$, and each pair of particles (edge)  with a value indicating their charge difference. The task is to predict the particles' locations after a fixed time. We follow the protocol of \citep{satorras2021n}. As baselines we use EGNN \citep{satorras2021n}, GNN \citep{gilmer2017neural}, TFN \citep{thomas2018tensor}, SE($3$)-Transformer \citep{fuchs2020se} and Radial Field \citep{2019equivariant}. We test our FA-GNN, as described in Section \ref{ss:graphs_perms_and_euclidean}. Table \ref{tab:n_body} logs the results, where the metric reported is the Mean Squared Error between predicted locations and ground truth. As can be seen, FA-GNN improves over the SOTA by more than 20\% . Note that the number of parameters used in FA-GNN and EGNN is roughly the same. More details are provided in  Appendix \ref{appendix:n-body}

\vspace{-5pt}
\section{Conclusions}\vspace{-5pt}
We present Frame Averaging, a generic and principled methodology for adapting existing (backbone) neural architectures to be invariant/equivariant to desired symmetries that appear in the data. We prove the method preserves the expressive power of the backbone model, and is efficient to compute in several cases of interest. We use FA to build universal GNNs, universal point cloud networks that are invariant/equivariant to Euclidean motions $E(3)$, and GNNs invariant/equivariant to Euclidean motions $E(3)$. We empirically validate the effectiveness of these models on several invariant/equivariant learning tasks. 
We believe the instantiations presented in the paper are only the first step in exploring the full potential of the FA framework, and there are many other symmetries and scenarios that can benefit from FA. For example, extending the invariance (or equivariance) of a model from a subgroup $H$ of $G$, to $G$. Further 
Interesting open questions, not answered by this paper are: 
What would be a way to systematically find efficient/small frames? How the frame choice effects the learning process? How to explore useful FA architectures and modules? 

\section*{Acknowledgments}
OP, MA and HB were supported by the European Research Council (ERC Consolidator Grant,
"LiftMatch" 771136) and also by a research
grant from the Carolito Stiftung (WAIC).



\bibliography{iclr2022_conference}

\begin{thebibliography}{71}
\providecommand{\natexlab}[1]{#1}
\providecommand{\url}[1]{\texttt{#1}}
\expandafter\ifx\csname urlstyle\endcsname\relax
  \providecommand{\doi}[1]{doi: #1}\else
  \providecommand{\doi}{doi: \begingroup \urlstyle{rm}\Url}\fi

\bibitem[Abboud et~al.(2021)Abboud, Ceylan, Grohe, and
  Lukasiewicz]{ACGL-IJCAI21}
Ralph Abboud, {\.I}smail~{\.I}lkan Ceylan, Martin Grohe, and Thomas
  Lukasiewicz.
\newblock The surprising power of graph neural networks with random node
  initialization.
\newblock In \emph{Proceedings of the Thirtieth International Joint Conference
  on Artifical Intelligence ({IJCAI})}, 2021.

\bibitem[Anderson et~al.(2019)Anderson, Hy, and Kondor]{anderson2019cormorant}
Brandon Anderson, Truong-Son Hy, and Risi Kondor.
\newblock Cormorant: Covariant molecular neural networks.
\newblock \emph{arXiv preprint arXiv:1906.04015}, 2019.

\bibitem[Ao et~al.(2021)Ao, Hu, Yang, Markham, and Guo]{ao2021spinnet}
Sheng Ao, Qingyong Hu, Bo~Yang, Andrew Markham, and Yulan Guo.
\newblock Spinnet: Learning a general surface descriptor for 3d point cloud
  registration.
\newblock In \emph{Proceedings of the IEEE/CVF Conference on Computer Vision
  and Pattern Recognition}, pp.\  11753--11762, 2021.

\bibitem[Atzmon et~al.(2018)Atzmon, Maron, and Lipman]{atzmon2018point}
Matan Atzmon, Haggai Maron, and Yaron Lipman.
\newblock Point convolutional neural networks by extension operators.
\newblock \emph{arXiv preprint arXiv:1803.10091}, 2018.

\bibitem[Balcilar et~al.(2021)Balcilar, H{\'e}roux, Ga{\"u}z{\`e}re, Vasseur,
  Adam, and Honeine]{balcilar2021breaking}
Muhammet Balcilar, Pierre H{\'e}roux, Benoit Ga{\"u}z{\`e}re, Pascal Vasseur,
  S{\'e}bastien Adam, and Paul Honeine.
\newblock Breaking the limits of message passing graph neural networks.
\newblock In \emph{Proceedings of the 38th International Conference on Machine
  Learning (ICML)}, 2021.

\bibitem[Battaglia et~al.(2018)Battaglia, Hamrick, Bapst, Sanchez-Gonzalez,
  Zambaldi, Malinowski, Tacchetti, Raposo, Santoro, Faulkner,
  et~al.]{battaglia2018relational}
Peter~W Battaglia, Jessica~B Hamrick, Victor Bapst, Alvaro Sanchez-Gonzalez,
  Vinicius Zambaldi, Mateusz Malinowski, Andrea Tacchetti, David Raposo, Adam
  Santoro, Ryan Faulkner, et~al.
\newblock Relational inductive biases, deep learning, and graph networks.
\newblock \emph{arXiv preprint arXiv:1806.01261}, 2018.

\bibitem[Batzner et~al.(2021)Batzner, Smidt, Sun, Mailoa, Kornbluth, Molinari,
  and Kozinsky]{batzner2021se}
Simon Batzner, Tess~E Smidt, Lixin Sun, Jonathan~P Mailoa, Mordechai Kornbluth,
  Nicola Molinari, and Boris Kozinsky.
\newblock Se (3)-equivariant graph neural networks for data-efficient and
  accurate interatomic potentials.
\newblock \emph{arXiv preprint arXiv:2101.03164}, 2021.

\bibitem[Bouritsas et~al.(2020)Bouritsas, Frasca, Zafeiriou, and
  Bronstein]{bouritsas2020improving}
Giorgos Bouritsas, Fabrizio Frasca, Stefanos Zafeiriou, and Michael~M
  Bronstein.
\newblock Improving graph neural network expressivity via subgraph isomorphism
  counting.
\newblock \emph{arXiv preprint arXiv:2006.09252}, 2020.

\bibitem[Breiding et~al.(2018)Breiding, Kozhasov, and
  Lerario]{breiding2018geometry}
Paul Breiding, Khazhgali Kozhasov, and Antonio Lerario.
\newblock On the geometry of the set of symmetric matrices with repeated
  eigenvalues.
\newblock \emph{Arnold Mathematical Journal}, 4\penalty0 (3):\penalty0
  423--443, 2018.

\bibitem[Chen et~al.(2019)Chen, Li, Xu, Chen, Wang, and
  Lin]{chen2019clusternet}
Chao Chen, Guanbin Li, Ruijia Xu, Tianshui Chen, Meng Wang, and Liang Lin.
\newblock Clusternet: Deep hierarchical cluster network with rigorously
  rotation-invariant representation for point cloud analysis.
\newblock In \emph{Proceedings of the IEEE/CVF Conference on Computer Vision
  and Pattern Recognition}, pp.\  4994--5002, 2019.

\bibitem[Chen et~al.(2021)Chen, Liu, Chen, Li, and Hill]{chen2021equivariant}
Haiwei Chen, Shichen Liu, Weikai Chen, Hao Li, and Randall Hill.
\newblock Equivariant point network for 3d point cloud analysis.
\newblock In \emph{Proceedings of the IEEE/CVF Conference on Computer Vision
  and Pattern Recognition}, pp.\  14514--14523, 2021.

\bibitem[Cohen et~al.(2019)Cohen, Weiler, Kicanaoglu, and
  Welling]{cohen2019gauge}
Taco Cohen, Maurice Weiler, Berkay Kicanaoglu, and Max Welling.
\newblock Gauge equivariant convolutional networks and the icosahedral cnn.
\newblock In \emph{International Conference on Machine Learning}, pp.\
  1321--1330. PMLR, 2019.

\bibitem[Cohen et~al.(2018)Cohen, Geiger, K{\"o}hler, and
  Welling]{cohen2018spherical}
Taco~S Cohen, Mario Geiger, Jonas K{\"o}hler, and Max Welling.
\newblock Spherical cnns.
\newblock In \emph{International Conference on Learning Representations}, 2018.

\bibitem[Dembo \& Zeitouni(2010)Dembo and Zeitouni]{DemboAmir2010LDTa}
Amir Dembo and Ofer Zeitouni.
\newblock \emph{Large Deviations Techniques and Applications}, volume~38 of
  \emph{Stochastic Modelling and Applied Probability}.
\newblock Springer Berlin / Heidelberg, Berlin/Heidelberg, 2010.
\newblock ISBN 9783642033100.

\bibitem[Deng et~al.(2021)Deng, Litany, Duan, Poulenard, Tagliasacchi, and
  Guibas]{deng2021vector}
Congyue Deng, Or~Litany, Yueqi Duan, Adrien Poulenard, Andrea Tagliasacchi, and
  Leonidas Guibas.
\newblock Vector neurons: A general framework for so (3)-equivariant networks.
\newblock \emph{arXiv preprint arXiv:2104.12229}, 2021.

\bibitem[Deng et~al.(2018)Deng, Birdal, and Ilic]{deng2018ppf}
Haowen Deng, Tolga Birdal, and Slobodan Ilic.
\newblock Ppf-foldnet: Unsupervised learning of rotation invariant 3d local
  descriptors.
\newblock In \emph{Proceedings of the European Conference on Computer Vision
  (ECCV)}, pp.\  602--618, 2018.

\bibitem[Dym \& Maron(2020)Dym and Maron]{dym2020universality}
Nadav Dym and Haggai Maron.
\newblock On the universality of rotation equivariant point cloud networks.
\newblock \emph{arXiv preprint arXiv:2010.02449}, 2020.

\bibitem[Esteves et~al.(2018)Esteves, Allen-Blanchette, Makadia, and
  Daniilidis]{esteves2018learning}
Carlos Esteves, Christine Allen-Blanchette, Ameesh Makadia, and Kostas
  Daniilidis.
\newblock Learning so (3) equivariant representations with spherical cnns.
\newblock In \emph{Proceedings of the European Conference on Computer Vision
  (ECCV)}, pp.\  52--68, 2018.

\bibitem[Finzi et~al.(2020)Finzi, Stanton, Izmailov, and
  Wilson]{finzi2020generalizing}
Marc Finzi, Samuel Stanton, Pavel Izmailov, and Andrew~Gordon Wilson.
\newblock Generalizing convolutional neural networks for equivariance to lie
  groups on arbitrary continuous data, 2020.

\bibitem[Fuchs et~al.(2020)Fuchs, Worrall, Fischer, and Welling]{fuchs2020se}
Fabian~B Fuchs, Daniel~E Worrall, Volker Fischer, and Max Welling.
\newblock Se (3)-transformers: 3d roto-translation equivariant attention
  networks.
\newblock \emph{arXiv preprint arXiv:2006.10503}, 2020.

\bibitem[Fulton \& Harris(2013)Fulton and Harris]{fulton2013representation}
William Fulton and Joe Harris.
\newblock \emph{Representation theory: a first course}, volume 129.
\newblock Springer Science \& Business Media, 2013.

\bibitem[F{\"u}rer(2010)]{furer2010power}
Martin F{\"u}rer.
\newblock On the power of combinatorial and spectral invariants.
\newblock \emph{Linear algebra and its applications}, 432\penalty0
  (9):\penalty0 2373--2380, 2010.

\bibitem[Gilmer et~al.(2017)Gilmer, Schoenholz, Riley, Vinyals, and
  Dahl]{gilmer2017neural}
Justin Gilmer, Samuel~S Schoenholz, Patrick~F Riley, Oriol Vinyals, and
  George~E Dahl.
\newblock Neural message passing for quantum chemistry.
\newblock In \emph{International conference on machine learning}, pp.\
  1263--1272. PMLR, 2017.

\bibitem[Gojcic et~al.(2019)Gojcic, Zhou, Wegner, and
  Wieser]{gojcic2019perfect}
Zan Gojcic, Caifa Zhou, Jan~D Wegner, and Andreas Wieser.
\newblock The perfect match: 3d point cloud matching with smoothed densities.
\newblock In \emph{Proceedings of the IEEE/CVF Conference on Computer Vision
  and Pattern Recognition}, pp.\  5545--5554, 2019.

\bibitem[Golub \& Van~Loan(1996)Golub and Van~Loan]{golub1996matrix}
Gene~H Golub and Charles~F Van~Loan.
\newblock Matrix computations. edition, 1996.

\bibitem[Hu et~al.(2021)Hu, Shuaibi, Das, Goyal, Sriram, Leskovec, Parikh, and
  Zitnick]{hu2021forcenet}
Weihua Hu, Muhammed Shuaibi, Abhishek Das, Siddharth Goyal, Anuroop Sriram,
  Jure Leskovec, Devi Parikh, and C~Lawrence Zitnick.
\newblock Forcenet: A graph neural network for large-scale quantum
  calculations.
\newblock \emph{arXiv preprint arXiv:2103.01436}, 2021.

\bibitem[Kingma \& Ba(2014)Kingma and Ba]{kingma2014adam}
Diederik~P Kingma and Jimmy Ba.
\newblock Adam: A method for stochastic optimization.
\newblock \emph{arXiv preprint arXiv:1412.6980}, 2014.

\bibitem[Kipf \& Welling(2016)Kipf and Welling]{kipf2016semi}
Thomas~N Kipf and Max Welling.
\newblock Semi-supervised classification with graph convolutional networks.
\newblock \emph{arXiv preprint arXiv:1609.02907}, 2016.

\bibitem[Klicpera et~al.(2021)Klicpera, Becker, and
  G{\"u}nnemann]{klicpera2021gemnet}
Johannes Klicpera, Florian Becker, and Stephan G{\"u}nnemann.
\newblock Gemnet: Universal directional graph neural networks for molecules.
\newblock \emph{arXiv preprint arXiv:2106.08903}, 2021.

\bibitem[Koch et~al.(2019)Koch, Matveev, Jiang, Williams, Artemov, Burnaev,
  Alexa, Zorin, and Panozzo]{Koch_2019_CVPR}
Sebastian Koch, Albert Matveev, Zhongshi Jiang, Francis Williams, Alexey
  Artemov, Evgeny Burnaev, Marc Alexa, Denis Zorin, and Daniele Panozzo.
\newblock Abc: A big cad model dataset for geometric deep learning.
\newblock In \emph{The IEEE Conference on Computer Vision and Pattern
  Recognition (CVPR)}, June 2019.

\bibitem[Kondor(2018)]{kondor2018n}
Risi Kondor.
\newblock N-body networks: a covariant hierarchical neural network architecture
  for learning atomic potentials.
\newblock \emph{arXiv preprint arXiv:1803.01588}, 2018.

\bibitem[Kondor et~al.(2018)Kondor, Son, Pan, Anderson, and
  Trivedi]{kondor2018covariant}
Risi Kondor, Hy~Truong Son, Horace Pan, Brandon Anderson, and Shubhendu
  Trivedi.
\newblock Covariant compositional networks for learning graphs.
\newblock \emph{arXiv preprint arXiv:1801.02144}, 2018.

\bibitem[Köhler et~al.(2019)Köhler, Klein, and Noé]{2019equivariant}
Jonas Köhler, Leon Klein, and Frank Noé.
\newblock Equivariant flows: sampling configurations for multi-body systems
  with symmetric energies, 2019.

\bibitem[LeCun et~al.(1998)LeCun, Bottou, Bengio, and
  Haffner]{lecun1998gradient}
Yann LeCun, L{\'e}on Bottou, Yoshua Bengio, and Patrick Haffner.
\newblock Gradient-based learning applied to document recognition.
\newblock \emph{Proceedings of the IEEE}, 86\penalty0 (11):\penalty0
  2278--2324, 1998.

\bibitem[Levin \& Peres(2017)Levin and Peres]{levin2017markov}
David~A Levin and Yuval Peres.
\newblock \emph{Markov chains and mixing times}, volume 107.
\newblock American Mathematical Soc., 2017.

\bibitem[Li et~al.(2019)Li, Bi, and Lee]{li2019discrete}
Jiaxin Li, Yingcai Bi, and Gim~Hee Lee.
\newblock Discrete rotation equivariance for point cloud recognition.
\newblock In \emph{2019 International Conference on Robotics and Automation
  (ICRA)}, pp.\  7269--7275. IEEE, 2019.

\bibitem[Li et~al.(2018)Li, Bu, Sun, Wu, Di, and Chen]{li2018pointcnn}
Yangyan Li, Rui Bu, Mingchao Sun, Wei Wu, Xinhan Di, and Baoquan Chen.
\newblock Pointcnn: Convolution on x-transformed points.
\newblock \emph{Advances in neural information processing systems},
  31:\penalty0 820--830, 2018.

\bibitem[Liu et~al.(2018)Liu, Yao, Choi, Sinha, and Ramani]{liu2018deep}
Min Liu, Fupin Yao, Chiho Choi, Ayan Sinha, and Karthik Ramani.
\newblock Deep learning 3d shapes using alt-az anisotropic 2-sphere
  convolution.
\newblock In \emph{International Conference on Learning Representations}, 2018.

\bibitem[Long et~al.(2015)Long, Shelhamer, and Darrell]{long2015fully}
Jonathan Long, Evan Shelhamer, and Trevor Darrell.
\newblock Fully convolutional networks for semantic segmentation.
\newblock In \emph{Proceedings of the IEEE conference on computer vision and
  pattern recognition}, pp.\  3431--3440, 2015.

\bibitem[Loukas(2020)]{Loukas2020What}
Andreas Loukas.
\newblock What graph neural networks cannot learn: depth vs width.
\newblock In \emph{International Conference on Learning Representations}, 2020.
\newblock URL \url{https://openreview.net/forum?id=B1l2bp4YwS}.

\bibitem[Maron et~al.(2018)Maron, Ben-Hamu, Shamir, and
  Lipman]{maron2018invariant}
Haggai Maron, Heli Ben-Hamu, Nadav Shamir, and Yaron Lipman.
\newblock Invariant and equivariant graph networks.
\newblock \emph{arXiv preprint arXiv:1812.09902}, 2018.

\bibitem[Maron et~al.(2019)Maron, Ben-Hamu, Serviansky, and
  Lipman]{maron2019provably}
Haggai Maron, Heli Ben-Hamu, Hadar Serviansky, and Yaron Lipman.
\newblock Provably powerful graph networks.
\newblock \emph{arXiv preprint arXiv:1905.11136}, 2019.

\bibitem[Morris et~al.(2019)Morris, Ritzert, Fey, Hamilton, Lenssen, Rattan,
  and Grohe]{morris2019weisfeiler}
Christopher Morris, Martin Ritzert, Matthias Fey, William~L Hamilton, Jan~Eric
  Lenssen, Gaurav Rattan, and Martin Grohe.
\newblock Weisfeiler and leman go neural: Higher-order graph neural networks.
\newblock In \emph{Proceedings of the AAAI Conference on Artificial
  Intelligence}, volume~33, pp.\  4602--4609, 2019.

\bibitem[Murphy et~al.(2019)Murphy, Srinivasan, Rao, and
  Ribeiro]{murphy2019relational}
Ryan Murphy, Balasubramaniam Srinivasan, Vinayak Rao, and Bruno Ribeiro.
\newblock Relational pooling for graph representations.
\newblock In \emph{International Conference on Machine Learning}, pp.\
  4663--4673. PMLR, 2019.

\bibitem[Murphy et~al.(2018)Murphy, Srinivasan, Rao, and
  Ribeiro]{murphy2018janossy}
Ryan~L Murphy, Balasubramaniam Srinivasan, Vinayak Rao, and Bruno Ribeiro.
\newblock Janossy pooling: Learning deep permutation-invariant functions for
  variable-size inputs.
\newblock \emph{arXiv preprint arXiv:1811.01900}, 2018.

\bibitem[Niepert et~al.(2016)Niepert, Ahmed, and Kutzkov]{niepert16}
Mathias Niepert, Mohamed Ahmed, and Konstantin Kutzkov.
\newblock Learning convolutional neural networks for graphs.
\newblock In Maria~Florina Balcan and Kilian~Q. Weinberger (eds.),
  \emph{Proceedings of The 33rd International Conference on Machine Learning},
  volume~48 of \emph{Proceedings of Machine Learning Research}, pp.\
  2014--2023, New York, New York, USA, 20--22 Jun 2016. PMLR.
\newblock URL \url{https://proceedings.mlr.press/v48/niepert16.html}.

\bibitem[Paszke et~al.(2019)Paszke, Gross, Massa, Lerer, Bradbury, Chanan,
  Killeen, Lin, Gimelshein, Antiga, et~al.]{paszke2019pytorch}
Adam Paszke, Sam Gross, Francisco Massa, Adam Lerer, James Bradbury, Gregory
  Chanan, Trevor Killeen, Zeming Lin, Natalia Gimelshein, Luca Antiga, et~al.
\newblock Pytorch: An imperative style, high-performance deep learning library.
\newblock \emph{Advances in neural information processing systems},
  32:\penalty0 8026--8037, 2019.

\bibitem[Pinkus(1999)]{pinkus1999approximation}
Allan Pinkus.
\newblock Approximation theory of the mlp model in neural networks.
\newblock \emph{Acta numerica}, 8:\penalty0 143--195, 1999.

\bibitem[Polyanskiy \& Wu(2014)Polyanskiy and Wu]{polyanskiy2014lecture}
Yury Polyanskiy and Yihong Wu.
\newblock Lecture notes on information theory.
\newblock \emph{Lecture Notes for ECE563 (UIUC) and}, 6\penalty0
  (2012-2016):\penalty0 7, 2014.

\bibitem[Puny et~al.(2020)Puny, Ben-Hamu, and Lipman]{puny2020global}
Omri Puny, Heli Ben-Hamu, and Yaron Lipman.
\newblock Global attention improves graph networks generalization, 2020.

\bibitem[Qi et~al.(2017{\natexlab{a}})Qi, Su, Mo, and Guibas]{qi2017pointnet}
Charles~R Qi, Hao Su, Kaichun Mo, and Leonidas~J Guibas.
\newblock Pointnet: Deep learning on point sets for 3d classification and
  segmentation.
\newblock In \emph{Proceedings of the IEEE conference on computer vision and
  pattern recognition}, pp.\  652--660, 2017{\natexlab{a}}.

\bibitem[Qi et~al.(2017{\natexlab{b}})Qi, Yi, Su, and Guibas]{qi2017pointnet++}
Charles~R Qi, Li~Yi, Hao Su, and Leonidas~J Guibas.
\newblock Pointnet++: Deep hierarchical feature learning on point sets in a
  metric space.
\newblock \emph{arXiv preprint arXiv:1706.02413}, 2017{\natexlab{b}}.

\bibitem[Romero \& Cordonnier(2021)Romero and Cordonnier]{romero2020group}
David~W Romero and Jean-Baptiste Cordonnier.
\newblock Group equivariant stand-alone self-attention for vision.
\newblock In \emph{ICLR}, 2021.

\bibitem[Sannai et~al.(2021)Sannai, Kawano, and Kumagai]{sannai2021equivariant}
Akiyoshi Sannai, Makoto Kawano, and Wataru Kumagai.
\newblock Equivariant and invariant reynolds networks, 2021.

\bibitem[Satorras et~al.(2021)Satorras, Hoogeboom, and Welling]{satorras2021n}
Victor~Garcia Satorras, Emiel Hoogeboom, and Max Welling.
\newblock E (n) equivariant graph neural networks.
\newblock \emph{arXiv preprint arXiv:2102.09844}, 2021.

\bibitem[Segol \& Lipman(2019)Segol and Lipman]{segol2019universal}
Nimrod Segol and Yaron Lipman.
\newblock On universal equivariant set networks.
\newblock \emph{arXiv preprint arXiv:1910.02421}, 2019.

\bibitem[Shuaibi et~al.(2021)Shuaibi, Kolluru, Das, Grover, Sriram, Ulissi, and
  Zitnick]{shuaibi2021rotation}
Muhammed Shuaibi, Adeesh Kolluru, Abhishek Das, Aditya Grover, Anuroop Sriram,
  Zachary Ulissi, and C~Lawrence Zitnick.
\newblock Rotation invariant graph neural networks using spin convolutions.
\newblock \emph{arXiv preprint arXiv:2106.09575}, 2021.

\bibitem[Tang et~al.(2019)Tang, Li, and Yu]{tang2019chebnet}
Shanshan Tang, Bo~Li, and Haijun Yu.
\newblock Chebnet: Efficient and stable constructions of deep neural networks
  with rectified power units using chebyshev approximations, 2019.

\bibitem[Thomas et~al.(2018)Thomas, Smidt, Kearnes, Yang, Li, Kohlhoff, and
  Riley]{thomas2018tensor}
Nathaniel Thomas, Tess Smidt, Steven Kearnes, Lusann Yang, Li~Li, Kai Kohlhoff,
  and Patrick Riley.
\newblock Tensor field networks: Rotation-and translation-equivariant neural
  networks for 3d point clouds.
\newblock \emph{arXiv preprint arXiv:1802.08219}, 2018.

\bibitem[Veličković et~al.(2018)Veličković, Cucurull, Casanova, Romero,
  Liò, and Bengio]{Petar2018graph}
Petar Veličković, Guillem Cucurull, Arantxa Casanova, Adriana Romero, Pietro
  Liò, and Yoshua Bengio.
\newblock Graph attention networks, 2018.

\bibitem[Wang et~al.(2018)Wang, Sun, Liu, Sarma, Bronstein, and
  Solomon]{wang2018dynamic}
Yue Wang, Yongbin Sun, Ziwei Liu, Sanjay~E Sarma, Michael~M Bronstein, and
  Justin~M Solomon.
\newblock Dynamic graph cnn for learning on point clouds.(2018).
\newblock \emph{arXiv preprint arXiv:1801.07829}, 2018.

\bibitem[Weiler et~al.(2018)Weiler, Geiger, Welling, Boomsma, and
  Cohen]{weiler20183d}
Maurice Weiler, Mario Geiger, Max Welling, Wouter Boomsma, and Taco Cohen.
\newblock 3d steerable cnns: Learning rotationally equivariant features in
  volumetric data.
\newblock \emph{arXiv preprint arXiv:1807.02547}, 2018.

\bibitem[Worrall \& Brostow(2018)Worrall and Brostow]{worrall2018cubenet}
Daniel Worrall and Gabriel Brostow.
\newblock Cubenet: Equivariance to 3d rotation and translation.
\newblock In \emph{Proceedings of the European Conference on Computer Vision
  (ECCV)}, pp.\  567--584, 2018.

\bibitem[Worrall et~al.(2017)Worrall, Garbin, Turmukhambetov, and
  Brostow]{worrall2017harmonic}
Daniel~E Worrall, Stephan~J Garbin, Daniyar Turmukhambetov, and Gabriel~J
  Brostow.
\newblock Harmonic networks: Deep translation and rotation equivariance.
\newblock In \emph{Proceedings of the IEEE Conference on Computer Vision and
  Pattern Recognition}, pp.\  5028--5037, 2017.

\bibitem[Xiao et~al.(2020)Xiao, Lin, Li, Geng, Chao, and
  Ding]{xiao2020endowing}
Zelin Xiao, Hongxin Lin, Renjie Li, Lishuai Geng, Hongyang Chao, and Shengyong
  Ding.
\newblock Endowing deep 3d models with rotation invariance based on principal
  component analysis.
\newblock In \emph{2020 IEEE International Conference on Multimedia and Expo
  (ICME)}, pp.\  1--6. IEEE, 2020.

\bibitem[Xu et~al.(2018{\natexlab{a}})Xu, Hu, Leskovec, and
  Jegelka]{xu2018powerful}
Keyulu Xu, Weihua Hu, Jure Leskovec, and Stefanie Jegelka.
\newblock How powerful are graph neural networks?
\newblock \emph{arXiv preprint arXiv:1810.00826}, 2018{\natexlab{a}}.

\bibitem[Xu et~al.(2018{\natexlab{b}})Xu, Fan, Xu, Zeng, and
  Qiao]{xu2018spidercnn}
Yifan Xu, Tianqi Fan, Mingye Xu, Long Zeng, and Yu~Qiao.
\newblock Spidercnn: Deep learning on point sets with parameterized
  convolutional filters.
\newblock In \emph{Proceedings of the European Conference on Computer Vision
  (ECCV)}, pp.\  87--102, 2018{\natexlab{b}}.

\bibitem[Yarotsky(2021)]{yarotsky2021universal}
Dmitry Yarotsky.
\newblock Universal approximations of invariant maps by neural networks.
\newblock \emph{Constructive Approximation}, pp.\  1--68, 2021.

\bibitem[Yu et~al.(2020)Yu, Wei, Tombari, and Sun]{yu2020deep}
Ruixuan Yu, Xin Wei, Federico Tombari, and Jian Sun.
\newblock Deep positional and relational feature learning for
  rotation-invariant point cloud analysis.
\newblock In \emph{European Conference on Computer Vision}, pp.\  217--233.
  Springer, 2020.

\bibitem[Zaheer et~al.(2017)Zaheer, Kottur, Ravanbakhsh, Poczos, Salakhutdinov,
  and Smola]{zaheer2017deep}
Manzil Zaheer, Satwik Kottur, Siamak Ravanbakhsh, Barnabas Poczos, Ruslan
  Salakhutdinov, and Alexander Smola.
\newblock Deep sets.
\newblock \emph{arXiv preprint arXiv:1703.06114}, 2017.

\bibitem[Zhang et~al.(2019)Zhang, Hua, Rosen, and Yeung]{zhang2019rotation}
Zhiyuan Zhang, Binh-Son Hua, David~W Rosen, and Sai-Kit Yeung.
\newblock Rotation invariant convolutions for 3d point clouds deep learning.
\newblock In \emph{2019 International Conference on 3D Vision (3DV)}, pp.\
  204--213. IEEE, 2019.

\end{thebibliography}
\bibliographystyle{iclr2022_conference}

\newpage
\appendix
\section{Proofs}\label{s:proofs}

\subsection{Proof of Theorem \ref{thm:frame}} \label{proof:frame}
\begin{proof}
First note that frame equivariance is defined to be $\gF(\rho_1(g)X) = g\gF(X)$ which in particular means $\abs{\gF(\rho_1(g)X)} = \abs{\gF(X)}$.

For invariance, let $g'\in G$
\begin{align*}
  \ip{\phi}_\gF (\rho_1(g')X) &= \frac{1}{|\gF(X)|}\sum_{g\in g'\gF(X)} \phi(\rho_1(g)^{-1} \rho_1(g') X)  = 
  \frac{1}{|\gF(X)|}\sum_{g\in \gF(X)}\phi(\rho_1(g'g)^{-1} \rho_1(g') X) \\
  &= \frac{1}{|\gF(X)|}\sum_{g\in \gF(X)} \phi(\rho_1(g)^{-1}X) = \ip{\phi}_\gF(X)
\end{align*}
and for equivariance
\begin{align*}
    \ip{\Phi}_\gF(\rho_1(g')X) &= \frac{1}{|\gF(X)|}\sum_{g\in g'\gF(X)} \rho_2(g) \Phi(\rho_1(g)^{-1} \rho_1(g') X) \\ &= 
    \frac{1}{|\gF(X)|}\sum_{g\in \gF(X)} \rho_2(g'g) \Phi(\rho_1(g'g)^{-1} \rho_1(g') X) \\
    &= \rho_2(g')\frac{1}{|\gF(X)|}\sum_{g\in \gF(X)} \rho_2(g) \Phi(\rho_1(g)^{-1} X) \\ &= \rho_2(g')\ip{\Phi}_\gF(X)
\end{align*}
\end{proof}


\subsection{Proof of Theorem \ref{thm:second_symmetry}} \label{proof:second_symmetry}
\begin{proof} 
The proof above \ref{proof:frame}, of Theorem \ref{thm:frame}, is a special case of the following where the second symmetry is chosen to be trivial. In principle the proofs are quite similar.

First note that frame equivariance and invariance, together with $\rho_1,\tau_1$ commuting mean  $$\gF(\gamma_1(h,g)X)=\gF(\tau_1(h)\rho_1(g)X)=\gF(\rho_1(g)\tau_1(h)X)=g \gF(\tau_1(h) X)=g\gF(X),$$
which in particular implies that $\abs{\gF(\gamma_1(h,g)X)}=\abs{\gF(X)}$. 
Let $(h',g')\in H\times G$ be arbitrary. Then,
\begin{align*}
  \ip{\phi}_\gF (\gamma_1(h',g')X) &= \frac{1}{|\gF(X)|}\sum_{g\in g'\gF(X)} \phi(\rho_1(g)^{-1} \tau_1(h')\rho_1(g') X)  \\ 
  &= \frac{1}{|\gF(X)|}\sum_{g\in \gF(X)} \phi(\rho_1(g'g)^{-1} \tau_1(h')\rho_1(g') X)  \\ 
  &= \frac{1}{|\gF(X)|}\sum_{g\in \gF(X)} \phi(\rho_1(g)^{-1} \tau_1(h') X)  \\ 
  &= 
  \frac{1}{|\gF(X)|}\sum_{g\in \gF(X)} \phi(\rho_1(g)^{-1}  X) \\
  & = \ip{\phi}_\gF(X)
\end{align*}
meaning that $\ip{\phi}_\gF$ is $H\times G$ invariant. Next,
\begin{align*}
    \ip{\Phi}_\gF(\gamma_1(h',g')X) &= \frac{1}{|\gF(X)|}\sum_{g\in g'\gF(X)} \rho_2(g) \Phi(\rho_1(g)^{-1} \tau_1(h')\rho_1(g') X) 
    \\ &= 
   \frac{1}{|\gF(X)|}\sum_{g\in \gF(X)} \rho_2(g'g) \Phi(\rho_1(g'g)^{-1} \tau_1(h')\rho_1(g') X)
   \\ &= 
   \frac{1}{|\gF(X)|}\sum_{g\in \gF(X)} \rho_2(g'g) \Phi(\rho_1(g)^{-1} \tau_1(h') X)
   \\ &= 
   \tau_2(h')\rho_2(g')\frac{1}{|\gF(X)|}\sum_{g\in \gF(X)} \rho_2(g) \Phi(\rho_1(g)^{-1}X)
   \\ &= 
   \gamma_2(h',g')\ip{\Phi}_\gF(X)
\end{align*}
showing that $\ip{\Phi}_\gF$ is $H\times G$ equivariant. 
\end{proof}

\subsection{Proof of Theorem \ref{thm:orbits_decomp_and_equality} and Corollary \ref{cor:prob}}
\label{a:orbits_decomp_and_equality}

\begin{proof}(Theorem \ref{thm:orbits_decomp_and_equality})
First we show $G_X$ acts on $\gF(X)$. For an arbitrary $h\in G_X$ and equivariant frame $\gF$, we have $\gF(X)=\gF(\rho_1(h)X)=h\gF(X)$, where in the first equality we used the fact that $h\in G_X$ and in the second the equivariance of $\gF$. This means that if $g\in\gF(X)$ and $h\in G_X$ then also $hg\in\gF(X)$, or in other words $\gF(X)$ is closed to actions from $G_X$. Furthermore, since a group acts on itself via bijections we have that the cardinality of all orbits is the same, $|[g]|=|G_X g|=|G_X|$, for all $[g]\in \gF(X)/G_X$. Lastly, the equivalence relation $g\sim h$ $\Longleftrightarrow$ $hg^{-1}\in G_X$ shows that $\gF(X)$ is a union of disjoint orbits of equal cardinality. 
\end{proof}
\begin{proof}(Corollary \ref{cor:prob})
Theorem \ref{thm:orbits_decomp_and_equality} asserts that all orbits $[g]\in\gF(X)/G_X$ have the same number of elements. Therefore, a random choice $g\in\gF(X)$ will have equal probability to land in each orbit.
\end{proof}

\subsection{The role of $m_\gF$ in approximation quality of \eqref{e:FA_approx}}\label{ss: role of m_F} 
To better understand the role of $m_{_\gF}$ in approximating \eqref{e:qoutient_FA} we first note that \eqref{e:FA_approx} can be written as $\ipp{\phi}_\gF(X)=\sum_{[g]\in \gF(X)/G_X} \hat{\mu}_{[g]} \phi(\rho_1(g_i)^{-1}X)$, where $\hat{\mu}$ is the empiricial distribution over $\gF(X)/G_X$, assigning to each element $[g]\in \gF(X)/G_X$ the fraction of samples landed in $[g]$, \ie, $\hat{\mu}_{[g]}=|\set{i\in [k]\vert g_i\in [g]}|k^{-1}$. We next present a lower bound on the probability of any particular $\hat{\mu}$ that provides a good approximation $\ipp{\phi}_\gF(X)\approx \ip{\phi}_{\gF}(X)$.
\begin{theorem}\label{thm:large_deviations}
Let $\gF$ be an equivariant frame. The probability of an arbitrary $\hat{\mu}\in \big\{\hat{\mu} \ \Big \vert \ \sup_{\phi\in\gQ}\abs{\ip{\phi}_\gF(X) - \ipp{\phi}_\gF(X)} \leq \eps \big\}$ is bounded from below as follows,
\begin{equation*}
    P(\hat{\mu})\geq (1+k)^{-{m_{_\gF}}}\exp(-2m_{_\gF}k \eps^2),
\end{equation*}
where $\gQ=\set{\phi\in\gC(V,\Real) \; \vert \; |\phi(X)|\leq 1, \forall X\in V}$ the set of bounded, continuous functions $V\too\Real$.
\end{theorem}
Before providing the proof let us note that this theorem provides a lower bound for each particular "good" empirical distribution $\hat{\mu}$. The main takeoff is that for fixed $k$ and $\eps$, the smaller $m_\gF$ the better the lower bound. The counter-intuitive behaviour of this bound w.r.t.~$k$ and $\eps$ stems from the fact that the size of the set of "good" $\hat{\mu}$, namely $\set{\phi\in\gC(V,\Real) \; \vert \; |\phi(X)|\leq 1, \forall X\in V}$ is increasing with $k$ and $\eps$.

\begin{proof} 
For brevity we denote $m=m_{_\gF}$. Our setting can be formulated as follows. We have the uniform probability distribution, denoted $\mu$, over the discrete space $[m]$ (representing the quotient set $\gF(X)/G_X$); $\mu_j=\frac{1}{m}$, $j\in [m]$, and we have numbers $a_j=\phi(\rho_1(h_j)^{-1}X)$, where $[h_j] \in \gF(X)/G_X$, $j\in [m]$ represent exactly one sample per orbit. In this notation $\ip{\phi}_\gF(X)=\sum_{j=1}^m \frac{a_j}{m}$, while the approximation in \eqref{e:FA_approx} takes the form $\ipp{\phi}_\gF(X)=\sum_{j=1}^m \hat{\mu}_j a_j$, where $\hat{\mu}_j=k_j/k$, where $k_j$ is the number of samples $g_i \in [h_j]$, $i\in [k]$, \ie, samples that landed in the $j$-th orbit. Therefore, 
$$\sup_{\phi\in \gQ}\abs{\ip{\phi}_\gF(X) - \ipp{\phi}_\gF(X)} = \sup_{|a_j|\leq 1} \abs{\sum_{j=1}^m a_j\parr{\frac{1}{m} - \hat{\mu}_j}} = \sum_{j=1}^m \abs{\frac{1}{m}-\hat{\mu}_j} = 2\norm{\mu-\hat{\mu}}_{\mathrm{TV}},$$
where the latter is the total variation norm for discrete measures \citep{levin2017markov}. We denote $H(\hat{\mu}|\mu)=\sum_{j=1}^m \log(\hat{\mu}_j)\log\parr{\frac{\hat{\mu}_j}{\mu_j}}$ the KL-divergence. The Pinsker inequality and its inverse for discrete positive measures are (see \eg, \citep{polyanskiy2014lecture}):
\begin{equation}\label{e:tv_H}
  \frac{1}{2}\norm{\hat{\mu}-\mu}_{\mathrm{TV}}^2 \leq  H(\hat{\mu}|\mu) \leq \frac{2}{\alpha} \norm{\hat{\mu}-\mu}_{\mathrm{TV}}^2,   
\end{equation}
where $\alpha = \min_{j\in [m]} \mu_j = m^{-1}$. 
Therefore, 
\begin{align*}
 \Gamma_\eps=\Big\{\hat{\mu} \ \Big \vert \ \sup_{\phi\in\gQ}\abs{\ip{\phi}_\gF(X) - \ipp{\phi}_\gF(X)}\leq \eps \Big\} = 
 \Big\{\hat{\mu} \ \Big \vert \ 2\norm{\hat{\mu}-\mu}_{\mathrm{TV}}\leq \eps \Big\}
 \subset 
 \Big\{\hat{\mu} \ \Big \vert \ H(\bar{\mu}|\mu) \leq \frac{\eps^2m}{2} \Big\}
\end{align*}
Now, application of Large Deviation Theory (Lemma 2.1.9 in \cite{DemboAmir2010LDTa}) provides that for $\hat{\mu}$ so that $H(\hat{\mu}|\mu)\leq \frac{\eps^2 m}{2}$:
$$P(\hat{\mu})\geq \frac{1}{(1+k)^m}e^{-\frac{km\eps^2}{2}}.$$

\end{proof}

\subsection{Proof of Theorem \ref{thm:maximally_expressive}}
\label{proof:maximally_expressive}

\begin{proof}
Let $\Psi\in \gC(V,W)$ be an arbitrary $G$ equivariant function, $\gF$ a bounded $G$ equivariant frame over a frame-finite domain $K$. Let $c>0$ be the constant from Definition \ref{def:frame}. For arbitrary $X\in K$,
\begin{align*}
    \norm{\Psi(X) - \ip{\Phi}_\gF(X)}_W 
    & 
    =
    \norm{\ip{\Psi}_\gF(X) - \ip{\Phi}_\gF(X)}_W 
    \\ 
    &
    \leq \frac{1}{|\gF(X)|}\sum_{g\in \gF(X)} \norm{\rho_2(g)  \Psi(\rho_1(g)^{-1}X) - \rho_2(g) \Phi(\rho_1(g)^{-1}X) }_W\\
    &
    \leq \max_{g\in \gF(X)}\norm{\rho_2(g)}_{\mathrm{op}} \norm{\Psi-\Phi}_{K_\gF,W}  \\
    & \leq c \norm{\Psi-\Phi}_{K_\gF,W}
\end{align*}
where in the first equality we used the fact that $\ip{\Psi}_\gF=\Psi$ since $\Psi$ is already equivariant. 
\end{proof}

\subsection{Proof of Proposition \ref{prop:frame_E3_equivariance}}
\begin{proof}
Let us prove $\gF$ is equivariant (\eqref{e:equi}). Consider a transformation $g=(\mR,\vt)\in G$, and let $(\mO,\vs)\in \gF(\mX)$, then 

$$\mC  =  \mO \Lambda\mO^T, \quad \vs = \frac{1}{n}\mX^T\one,$$
where $\mO\Lambda\mO^T$ is the eigen decomposition of $C$.
Note that the group product of these transformations is
$$(\mR,\vt)(\mO,\vs) = (\mR\mO, \mR\vs+\vt).$$
We need to show $(\mR\mO,\mR\vs+\vt)\in\gF\parr{ (\mR,\vt)\mX}$. Indeed, $\mR\mO\in O(3)$ and consists of eigenvectors of $\mC=\mR \mX^T(\mI-\frac{1}{n}\one\one^T)\mX\mR^T$ as can be verified with a direct computation. If $\mO,\mR\in SO(d)$ then also $\mR\mO\in SO(d)$.   Furthermore $$\frac{1}{n}\parr{\mX\mR^T + \one\vt^T}^T\one = \frac{1}{n}\parr{\mR\mX^T + \vt\one^T}\one = \mR\vs+\vt$$ as required. We have shown $(\mR,\vt)\gF(\mX)\subset \gF((\mR,\vt)\mX)$ for all $\mX$ and $(\mR,\vt)$. To show the other inclusion let $\mX=(\mR,\vt)^{-1}\mY$ and get $\gF((\mR,\vt)^{-1}\mY)\subset (\mR,\vt)^{-1}\gF(\mY)$ that also holds for all $\mY$ and $(\mR,\vt)$. In particular $(\mR,\vt)\gF(\mX)\supset \gF((\mR,\vt)\mX)$. The frame $\gF$ is bounded since for compact $K\subset \Real^{n\times d}$, the translations $n^{-1}\mX^T\one$ are compact and therefore uniformly bounded for $X\in K$, and orthogonal matrices always satisfy $\norm{\mR}_2 = 1$. 
\end{proof}

\subsection{Proof of Proposition \ref{prop:sorting_frame}}\label{proof:sorting_frame}

\begin{proof}
First note that by definition $\mS(\tX)$ is equivariant in rows, namely $\mS(\rho_1(g)\tX)=\rho_1(g)\mS(\tX)$. Therefore if $g\in \gF(\tX) \subset S_n$, then by definition of the frame $\rho_1(g)\mS$ is sorted. Therefore, $\rho_1(g)\mS=\rho_1(gh) \rho_1(h)^{-1} \mS(\tX)=\rho_1(gh)\mS(\rho_1(h)^{-1}\tX)$ is sorted and we get that $gh\in \gF(\rho_1(h)^{-1}\tX)$. We proved $\gF(\tX)h \subset \gF(\rho_1(h)^{-1}\tX)$  for all $h\in G$ and all $\tX\in V$. Taking $\tX=\rho_1(h)\tY$ for an arbitrary $\tY\in V$, and $h=g^{-1}$ for arbitrary $g\in G$, we get $\gF(\rho_1(g^{-1})\tY)\subset \gF(\tY)g$, for all $g\in G$ and $\tY\in V$. We proved  $\gF(\tX)h = \gF(\rho_1(h)^{-1}\tX)$, which amounts to right action equivariance, see \eqref{e:frame_right}. The frame is bounded since $\rho_1(G)$ is a finite set. 
\end{proof}

\section{Empirical frame analysis}\label{s:frame_analysis}

\paragraph{Repeating eigenvalues.}

\begin{wrapfigure}[15]{r}{0.4\textwidth}
  \vspace{-15pt}
   \includegraphics[width=0.4\textwidth]{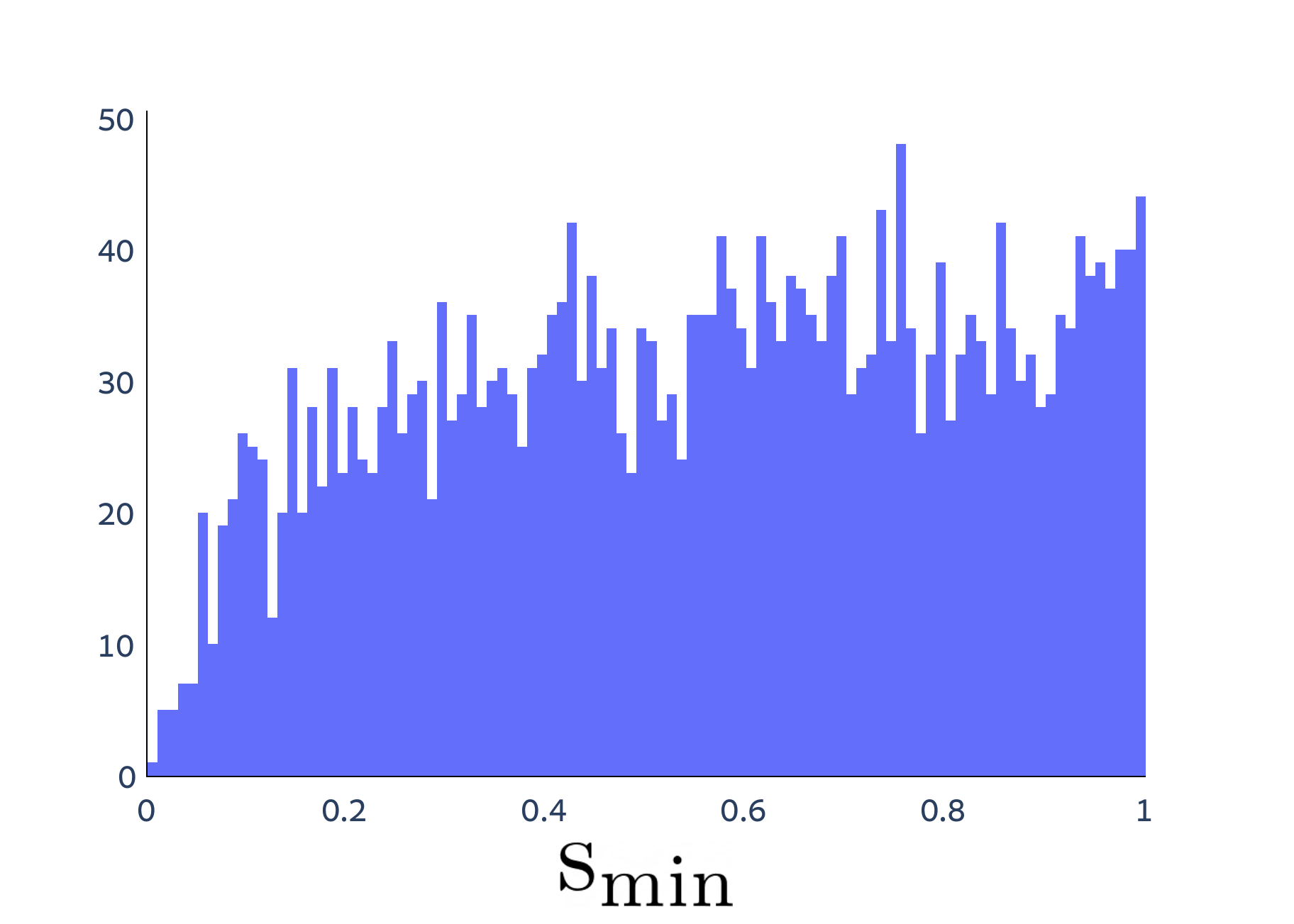}
  \vspace{-15pt}
  \caption{Minimal spacing histogram over the training set of $n$-body dataset \citep{satorras2021n} .}\label{fig:spacing}
\end{wrapfigure}
We empirically test the likeliness of repeating eigenvalues in the covariance matrix used from the definition of frame in Section \ref{ss:point_clouds}. We use the data from the $n$-body dataset \citep{satorras2021n}. Let $\mX\in\Real^{n\times 3}$ represent the set of particles locations (each set centered around $0$ and scaled to have $\max_{x_i\in\mX}\norm{x_i}_2=1$) and $\lambda_1\leq\lambda_2\leq\lambda_3$ be the eigenvalues, sorted in increasing order, of the covariance matrix $\mC=(\mX-\one\vt^T)^T(\mX-\one\vt^T)$ . In order to measure the proximity of eigenvalues across the dataset we use the notion of eigenvalues spacing $s_i = \frac{\lambda_{i+1}-\lambda_i}{\bar{s}}$, $i=1,2$, 
where $\bar{s}=\frac{\lambda_3-\lambda_1}{2}$ is the mean spacing. Furthermore we define $s_{min}=\min\set{s_1,s_2}$ as the minimal normalized spacing, a ratio that indicates how close the spectrum is to having repeating eigenvalues. Figure \ref{fig:spacing} presents a histogram of the minimal spacing over the training set of the $n$-body dataset (consists of $3000$ particles sets). The minimal spacing encountered in this experiment is of order $10^{-2}$. This empirically justifies the usage of finite frames for $E(d)$ equivariance.

\begin{wrapfigure}[15]{r}{0.4\textwidth}
  \begin{center}
  \vspace{-35pt}
    \includegraphics[width=0.4\textwidth]{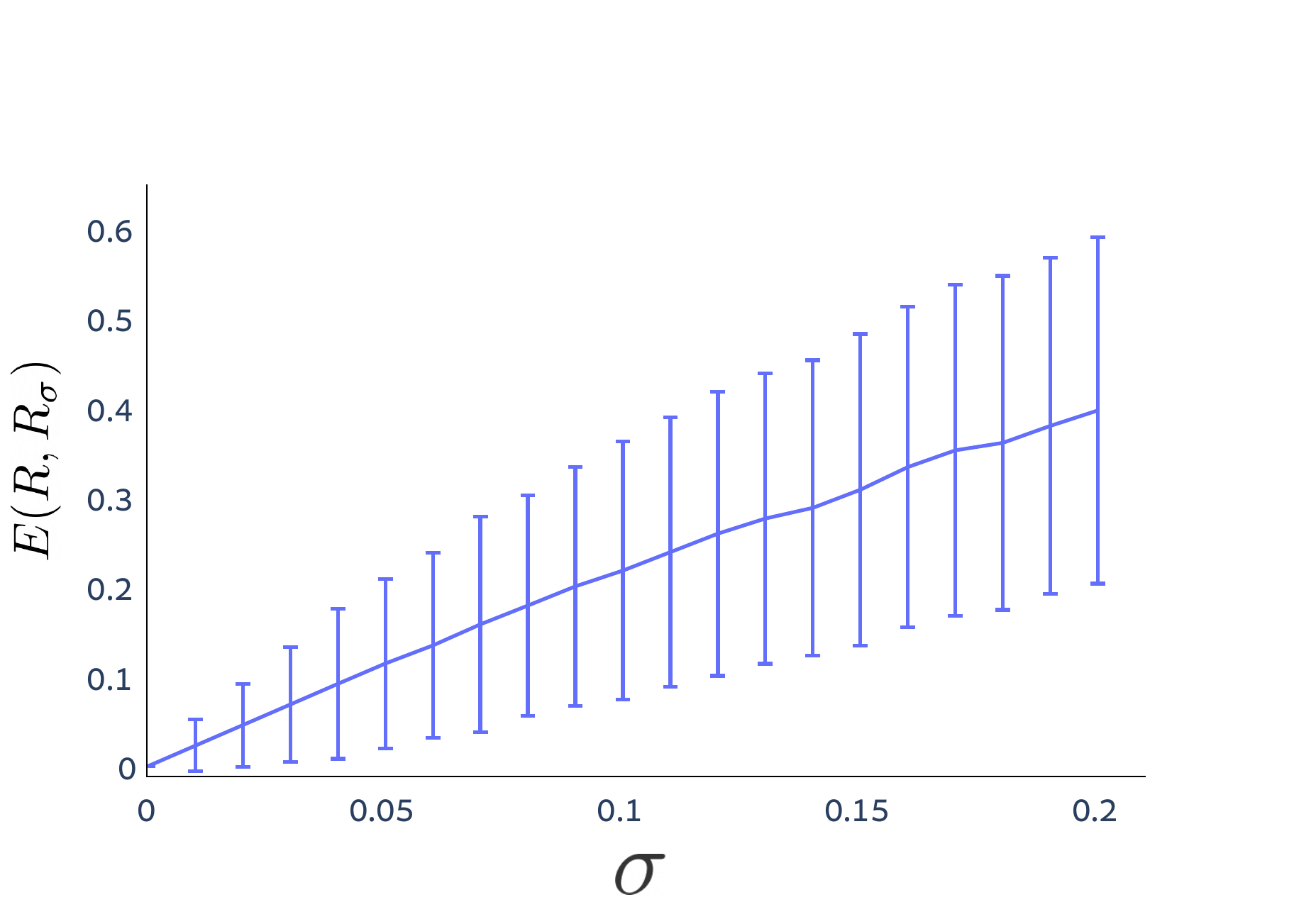}
  \end{center}
  \vspace{-15pt}
  \caption{Distance between original and noisy frames as a function of $\sigma$ (we plot average and std). The result is reported over the training set of $n$-body dataset \citep{satorras2021n} .}\label{fig:distance}
\end{wrapfigure}

\paragraph{Frame stability.} Here we test stability of our $O(d)$ frame defined in Section \ref{ss:point_clouds}. By stability we mean the magnitude of change of a frame w.r.t.~the change in the input $\mX$. A desired attribute of the constructed frame is to be stable, that is, to exhibit small changes if the input is perturbed. 
We quantify this stability by comparing the distance between our frames to frames constructed by a noisy input. As in the previous experiment used the data from the $n$-body dataset \citep{satorras2021n}. Let $\mX\in\Real^{n\times 3}$ represent the set of particles locations (each set centered around $0$ and scaled to have $\max_{x_i\in\mX}\norm{x_i}_2=1$) and $\mX_{\sigma} = \mX + \mZ$ where $z_{i,j}\sim\mathcal{N}(0,\sigma)$ is the noisy input sample . We compute $\gF(\mX)$ and $\gF(\mX_{\sigma})$ and choose  representatives from each set $g=(\mR,\vt)\in\gF(\mX),g_{\sigma}=(\mR_\sigma,\vt_\sigma)\in\gF(\mX_{\sigma})$. We measure the distance between the frames as a function of the representatives -
$$E(\mR,\mR_{\sigma})=\frac{1}{3}\sum_{i=1}^3\sqrt{1-\langle \mR_{:,i},(\mR_{\sigma})_{:,i}\rangle^2}$$
Notice that in the case of simple spectrum the distance is invariant to the selection of representatives. In Figure \ref{fig:distance} we plot distance of original and noisy frames (and its standard deviation) as function of noise level $\sigma$. The plot validates the continuity of frames in the simple spectrum case.

\begin{figure}[t]
\centering
     \includegraphics[width=0.8\columnwidth]{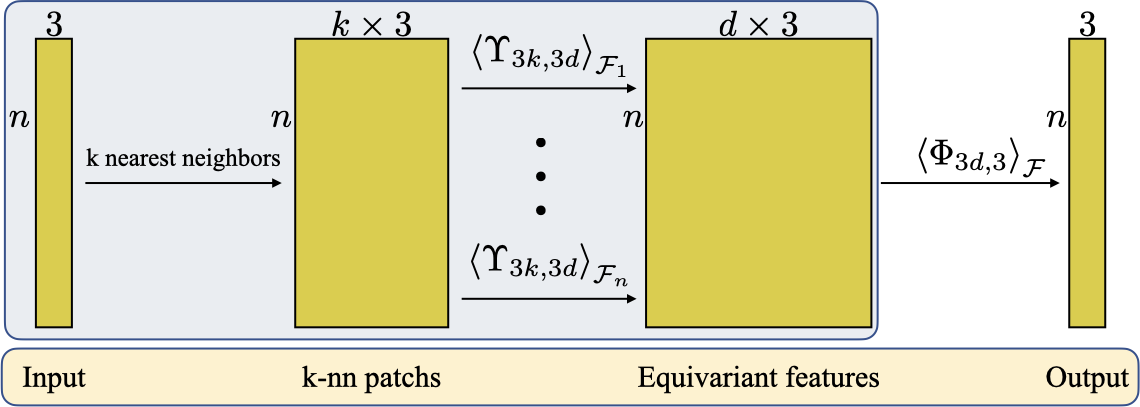}
    \caption{FA-Local-PointNet architecture.}
    \label{fig:falocal}
    
\end{figure}

\section{Implementation Details}
\label{as:id}
\subsection{Point Clouds: Normal Estimation} \label{app:impl_dets_PC}
Here we provide implementation details for the experiment in section \ref{ss:normal_estimation}. 
\paragraph{PointNet architecture.}
Our backbone PointNet is based on the object part segmentation network from \cite{qi2017pointnet}. The network consists of layers of the form 
\begin{align*}
\mathrm{FC}(n,d_{\text{in}},d_{\text{out}}):\mX &\mapsto \nu\parr{\mX \mW + \one \vb^T } \\
\mathrm{MaxPool}(n,d_{\text{in}}) : \mX &\mapsto \one [\max{\mX \ve_i}]
\end{align*}
where $\mX \in \Real ^ {n\times d_\text{in}}$, $\mW\in \Real^{d_\text{in}\times d_\text{out}}$, $\vb\in\Real^{d_\text{out}}$ are the learnable parameters, $\one\in\Real^n$ is the vector of all ones, $[\cdot]$ is the concatenation operator, $\ve_i$ is the standard basis in $\Real^{d_\text{in}}$, and $\nu$ is the $\mathrm{ReLU}$ activation. In this experiement, for the PointNet baseline and for $\Phi_{3,3}$ (the backbone in FA-PointNet), we used the following architecture:
\begin{align*}
&\mathrm{FC}(512,3,64) \stackrel{L_1}\rightarrow \mathrm{FC}(512,64,128)  \stackrel{L_2}\rightarrow \mathrm{FC}(512,128,128) \stackrel{L_3}\rightarrow  \mathrm{FC}(512,128,512) \stackrel{L_4}\rightarrow \\
&\mathrm{FC}(512,512,2048) \stackrel{L_5} \rightarrow \mathrm{MaxPool}(512,2048)  \stackrel{L_6}\rightarrow  [L_1,L_2,L_3,L_4,L_5,L_6] \stackrel{L_7} \rightarrow   \\
&\mathrm{FC}(512,4028,256) \stackrel{L_8}\rightarrow \mathrm{FC}(512,256,128)  \stackrel{L_9} \rightarrow \mathrm{FC}(512,128,3).
\end{align*}
Note that the original PointNet network also contains two T-Net networks, applied to the input and to $L_3$ (the output of the third layer). Similarly, our baseline implementation made a use of the same T-Net networks. Note that the T-Net networks were \emph{not} part of our FA-PointNet backbone architectures $\Phi_{3,3}$.

The FA-Local-Pointnet architecture can be seen as a composition of two parts (see Figure \ref{fig:falocal}). The first part outputs equivariant features by applying the same MLP backbone $\Upsilon_{3k,3d}$ with FA on each point's k-nn patch. Then, from each point's equivariant features, the second part of the network outputs an equivariant normal estimation. Note that the second part, $\ip{\Phi}_{\gF}$, is an FA-Pointnet network applied to a point cloud of dimensions $\Real ^{n \times 3d}$.
For FA-Local-PointNet we made the following design choices. Each point patch is constructed as its $k$ nearest neighbors, with $k=20$, $\mX_i \in \Real ^{20\times 3}$. Then, the backbone $\Upsilon_{3*20,3*42}$ is applied to all patches, with the following PointNet layers:
\begin{align*}
&\mathrm{FC}(512,3*20,3*21) \rightarrow \mathrm{FC}(512,3*21,3*42)\rightarrow \mathrm{FC}(512,3*42,3*42).
\end{align*}
The second backbone, $\Phi_{3*42,3}$, is built from the following PointNet layers:
\begin{align*}
&\mathrm{FC}(512,3*32,128) \stackrel{L_1}\rightarrow \mathrm{FC}(512,128,256)  \stackrel{L_2}\rightarrow \mathrm{FC}(512,256,256) \stackrel{L_3}\rightarrow  \mathrm{FC}(512,256,512) \stackrel{L_4}\rightarrow \\
&\mathrm{FC}(512,512,2048) \stackrel{L_5} \rightarrow \mathrm{MaxPool}(512,2048)  \stackrel{L_6}\rightarrow  [L_1,L_2,L_3,L_4,L_5,L_6] \stackrel{L_7} \rightarrow   \\
&\mathrm{FC}(512,5248,256) \stackrel{L_8}\rightarrow \mathrm{FC}(512,256,128)  \stackrel{L_9} \rightarrow \mathrm{FC}(512,128,3).
\end{align*}

\paragraph{DGCNN architecture.}
Our backbone DGCNN architecture, $\Phi_{3,3}$, is based on the object part segmentation network from \cite{wang2018dynamic}. It consists of $\mathrm{EdgeConv}(n,d_{\text{in}},d_{\text{out}})$, $\mathrm{FC}(n,d_{\text{in}},d_{\text{out}})$, and $\mathrm{MaxPool}$ layers.
\begin{align*}
&\mathrm{EdgeConv}(512,3,64) \stackrel{L_1}\rightarrow \mathrm{EdgeConv}(512,64,64)  \stackrel{L_2}\rightarrow \mathrm{EdgeConv}(512,64,64) \stackrel{L_3}\rightarrow  \\
&\mathrm{FC}(512,64,1024) \stackrel{L_4}\rightarrow \mathrm{MaxPool}(512,1024)  \stackrel{L_5}\rightarrow  [L_1,L_2,L_3,L_4,L_5] \stackrel{L_6} \rightarrow   \\
&\mathrm{FC}(512,1216,256) \stackrel{L_7}\rightarrow \mathrm{FC}(512,256,128)  \stackrel{L_8} \rightarrow \mathrm{FC}(512,128,3).
\end{align*}
Note that the DGCNN architecture incorporates T-Net network applied to the input.

\paragraph{Training details.}
We trained our networks using the \textsc{Adam} \citep{kingma2014adam} optimizer, setting the batch size to $32$ and $16$ for PointNet and DGCNN respectively. We set a fixed learning rate of $0.001$. All models were trained for $250$ epochs. Training was done on a single Nvidia V-100 GPU, using \textsc{pytorch} deep learning framework \citep{paszke2019pytorch}.


\subsection{Graphs: Expressive Power}\label{appendix:expressive power}
We provide implementation details for the experiments in \ref{ss:expressive power experiment}. We used two different universal backbones, MLP and GIN equipped with identifiers. The details of those architectures are presented here.
\paragraph{FA/GA GIN+ID architecture.} 
The GIN+ID backbone is based on the GIN \citep{xu2018powerful} network with an addition of identifiers as node features in order to increase expressiveness. For the experiments we used a three-layer GIN with a feature dimension of size 64 and a $\mathrm{ReLU}$ activation function. For the added identifiers we defined the input node features as $\mX^0 = [\mX^0,\mI_n]$, where $n$ is the number of nodes in the graph, and to handle graphs of different sizes (EXP dataset) we padded the node features with zeros to fit the size of the maximal graph. Note that we did not apply the permutation generated by the frame on the identifiers.

\paragraph{FA/GA MLP architecture.} 
We used two different MLP networks for the EXP dataset and the GRAPH8c, due to the different size of graphs in the datasets. In the EXP dataset the maximal graph size is $64$ and every node in the graph has a one dimensional binary feature, therefore the input for the MLP network is a flatten representation of the graph (with additional padding according to the graph size) $\vx\in\R^{64^2+ 64}$. Our architecture consists of layers of the form
\begin{align*}
\mathrm{FC}(d_{\text{in}},d_{\text{out}}):\vx &\mapsto \nu\parr{ \mW\vx + \vb } 
\end{align*}
where $\mW\in \Real^{d_{\text{out}}\times d_{\text{in}}}$, $\vb\in\Real^{d_{\text{out}}}$ are the learnable parameters. The final output of the network is denoted by $(d_{out})$ where $d_{out}$ is the output dimension.

The MLP network structure for the EXP-classify task: 
\begin{align*}
&\mathrm{FC}(4160, 2048) \stackrel{L_1}\rightarrow \mathrm{FC}(2048, 4096)  \stackrel{L_2}\rightarrow \mathrm{FC}(4096, 2048) \stackrel{L_3}\rightarrow \\ &\mathrm{FC}(2048, 10) \stackrel{L_4}\rightarrow 
\mathrm{FC}(10, 1) \stackrel{L_5}\rightarrow (1)
\end{align*}
with $\mathrm{ReLU}$ as the activation function. For the EXP task, which has $10$-dimensional output for each graph we omitted the last layer.
The GRAPH8c is composed of all the non-isomorphic connected graphs of size $8$, hence we did not used any padding of the input here. The nodes have no features and we just used a flatten version of the adjacency matrix. The architecture for the GRAPH8c task:
\begin{align*}
&\mathrm{FC}(64,128) \stackrel{L_1}\rightarrow \mathrm{FC}(128,64)  \stackrel{L_2}\rightarrow \mathrm{FC}(64,10) \stackrel{L_3}\rightarrow (10)
\end{align*}
\paragraph{Training details.}
We followed the protocol from \citep{balcilar2021breaking} and trained our model with batch size $100$ for 200 epochs. The learning rate was set to $0.001$ and did not change during training. For optimization we used the \textsc{Adam} optimizer. Training was done on a single Nvidia RTX-$8000$ GPU, using \textsc{pytorch} deep learning framework.

\paragraph{Invariance Evaluation.}
We quantify the permutation invariance of a model $\phi$ by comparing output of randomly permuted graphs $\rho_1(g_i)\tX$, $g_i\in S_n$,  $i\in[50]$. The evaluation metric used is the \emph{invariance error}, defined by 
 $$\frac{1}{m}\sum_{i=1}^m \norm{\phi(\rho_1(g_i) \tX) - \vv}_2 ,$$
 where $\vv = \frac{1}{m}\sum_{i=1}^m \phi(\rho_1(g_i)\tX)$. 

The permutation \emph{invariance error} of the models (FA/GA) was measured as a function of the sample size $k$. We iterated over the entire GRAPH8c dataset and for every graph computed the \emph{invariance error} for the FA-MLP ,GA-MLP and regular MLP models (all with the exact same backbone network). The results presented in Figure \ref{fig:inverr_m_f} (left) are normalized by the error of 
the regular MLP model. The backbone MLP we used for this experiment is of the form:
\begin{align*}
&\mathrm{FC}(64,128) \stackrel{L_1}\rightarrow \mathrm{FC}(128,128)  \stackrel{L_2}\rightarrow 
\mathrm{FC}(128,128)   \stackrel{L_3}\rightarrow  
\mathrm{FC}(128,10)   \stackrel{L_4}\rightarrow  (10)
\end{align*}

\subsection{Graphs: $n$-Body Problem}\label{appendix:n-body}
This section describes the implementation details for the $n$-body experiment from Section \ref{ss:n-body}. 

\paragraph{FA-GNN architecture.}
We use the GNN architecture of \cite{gilmer2017neural} our base GNN layer, where for each node $i\in [n]$ the update rule follows,
\begin{equation*}
    \vm_{ij} = \phi_e(\vh_i^l,\vh_j^l,a_{ij})
\end{equation*}
\begin{equation*}
    \vm_i = \sum_{j\in\mathcal{N}(i)}\vm_{ij}
\end{equation*}
\begin{equation*}
    \vh^{l+1}_i= \phi_h(\vh^l_i,\vm_i)
\end{equation*}
where $\gN(i)$ are the indices of neighbors of vertex $i$, $\vh_i^l$ is the embedding of node $i$ at layer $l$ and $a_{ij}\in\set{-1,1}\times \Real^+$ are edge attributes representing attraction or repelling between pairs of particles and their distance. $\vh^0\in\R^{n\times 6}$ represents the nodes input features, which in this experiment are a concatenation of the nodes initial $3$D position (rotation and translation equivariant) and velocities (rotation-equivariant and translation-invariant). We used node feature dimension of size $60$ to maintain a fair comparison with the baselines \citep{satorras2021n}. The functions $\phi_e$ and $\phi_h$ are implemented as a two-layer MLP with the $\mathrm{SiLU}$ activation function (also chosen for consistency purposes) with an hidden dimension of $121$ and $120$ respectively.
To maintain fair comparison with \citep{satorras2021n} our network is composed of $4$ GNN layers. $\Phi^{{\scriptscriptstyle (1)}}_{6,3d'} $ has an additional linear embedding of the features as a prefix to the GNN layers while $\Phi^{{\scriptscriptstyle (4)}}_{3d',3} $ is equipped with a two-layer MLP ($\mathrm{SiLU}$ activation) as a decoder to extract final positions. 


\paragraph{Training details.}
We followed the protocol from \citep{satorras2021n} and trained our model with batch size $100$ for $10000$ epochs. The learning rate was set to $0.001$ and did not changed during training. For optimization we used the \textsc{Adam} optimizer. Training was done on a single Nvidia RTX-$6000$ GPU, using \textsc{pytorch} deep learning framework.

\end{document}


\maketitle
\section{Proofs}

\section{Implementation details}
\subsection{Point Cloud: Normal Estimation}
\subsection{Point Cloud: $n$-body problem}
\subsection{Graph: Molecular data - QM9}
\subsection{Graph: Expressive Power}